\documentclass[runningheads]{llncs}
\usepackage{graphicx}

\usepackage{amsmath,amssymb} 
\usepackage{color}
\usepackage{booktabs}
\usepackage{multirow}
\usepackage{subfig,floatrow}
\usepackage{microtype}
\usepackage[font=it]{caption}
\newfloatcommand{capbtabbox}{table}[][\FBwidth]

\usepackage{tikz}
\def\labelledpic#1#2{
\begin{tikzpicture}
\node (pic) {#1};
\path[fill=white,draw=gray,thick] (pic.south west) +(3ex,3ex) circle (2ex)
   node {#2};
\end{tikzpicture}
}

\def\figref#1{Fig.~\ref{fig:#1}}

\let\subparagraph\paragraph
\usepackage[compact]{titlesec}

\begin{document}

\title{Creatures great and SMAL: Recovering the shape and motion of animals from video\thanks{The authors would like to thank GlaxoSmithKline for sponsoring this work.}} 
\titlerunning{Creatures great and SMAL} 



\author{Benjamin Biggs\inst{1} \and
Thomas Roddick\inst{1} \and
Andrew Fitzgibbon\inst{2} \and
Roberto Cipolla\inst{1}}

%


\authorrunning{B. Biggs et al.} 


\institute{University of Cambridge, Department of Engineering, Trumpington Street, Cambridge, CB2 1PZ, UK \email{\{bjb56, tr346, rc10001\}@cam.ac.uk}\and
Microsoft Research, 21 Station Road, Cambridge, CB1 2FB, UK \email{awf@microsoft.com}}
\maketitle

\begin{abstract}
We present a system to recover the 3D shape and motion of a wide variety of quadrupeds from video.  The system comprises a machine learning front-end which predicts candidate 2D joint positions, a discrete optimization which finds kinematically plausible joint correspondences, 
and an energy minimization stage which fits a detailed 3D model to the image. In order to overcome the limited availability of motion capture training data from animals, and the difficulty of generating realistic synthetic training images, the system is designed to work on silhouette data.  The joint candidate predictor is trained on synthetically generated silhouette images, and at test time, deep learning methods or standard video segmentation tools are used to extract silhouettes from real data. The system is tested on animal videos from several species, and shows accurate reconstructions of 3D shape and pose.
\end{abstract}
\section{Introduction}

Animal welfare is an important concern for business and society, with an estimated 70 billion animals currently living under human care~\cite{FAOSTAT}. Monitoring and assessment of animal health can be assisted by obtaining accurate measurements of an individual's shape, volume and movement. These measurements should be taken without interfering with the animal's normal activity, and are needed around the clock, under a variety of lighting and weather conditions, perhaps at long range (e.g.\ in farm fields or wildlife parks). Therefore a very wide range of cameras and imaging modalities must be handled. For small animals in captivity, a depth camera might be possible, but techniques which can operate solely from intensity data will have a much wider range of applicability.

We address this problem using techniques from the recent human body and hand tracking literature, combining machine learning and 3D model fitting.  A discriminative front-end uses a deep hourglass network to identify candidate 2D joint positions. These joint positions are then linked into coherent skeletons by solving an optimal joint assignment problem, and the resulting skeletons create an initial estimate for a generative model-fitting back-end to yield detailed shape and pose for each frame of the video.  

\def\bb{\rule{2in}{0pt}\rule{0pt}{1in}}

\begin{figure}[t]
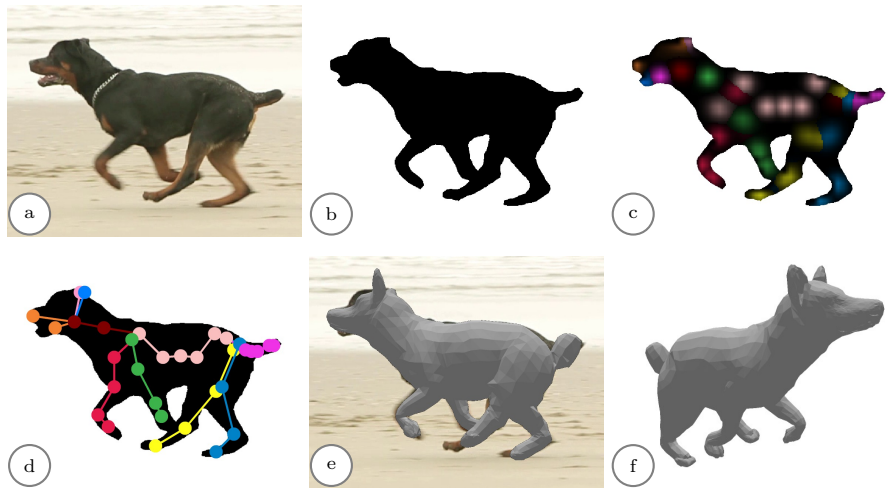

\def\p#1#2{\parbox{0.32\linewidth}{
\labelledpic{\includegraphics[trim={4cm 7cm 4cm 7cm},clip,width=\linewidth]{sys_overview_pred_2/#2.jpg}}
{\scriptsize #1}}}
\def\ps#1#2{\parbox{0.32\linewidth}{
\labelledpic{\includegraphics[trim={0cm 2cm 0cm 2cm},clip,width=\linewidth]{sys_overview_pred_2/#2.jpg}}
{\scriptsize #1}}}
\p a{rgb}             \p b{target}         \p c{heatmap}
\p d{skeleton_sil}    \p e{00048_overlay}  \ps f{00048_alternative}
\caption{{\bf System overview}: input video (a) is automatically processed using DeepLabv3+~\cite{deeplabv3plus} to produce silhouettes (b), from which 2D joint predictions are regressed in the form of heatmaps (c).  Optimal joint assignment (OJA) finds kinematically coherent 2D-to-3D correspondences~(d), 
which initialize a 3D shape model, optimized to match the silhouette~(e). 
Alternative view shown in (f).
}
\label{fig:overview}
\end{figure}

Although superficially similar to human tracking, animal tracking (AT) has some interesting differences that make it worthy of study:

\subsubsection*{Variability.}
In one sense, AT is simpler than human tracking as animals generally do not wear clothing. However, variations in surface texture are still considerable between individuals, and the variety of shape across and within species is considerably greater.  If tracking is specialized to a particular species, then shape variation is smaller, but training data is even harder to obtain.

\subsubsection*{Training data.}
For human tracking, hand labelled sequences of 2D segmentations and joint positions have been collected from a wide variety of sources~\cite{andriluka14cvpr,mscoco,johnson2010clustered}. Of these two classes of labelling, animal {\em segmentation} data is available in datasets such as MSCOCO~\cite{mscoco}, PASCAL VOC~\cite{pascal-voc-2012} and DAVIS~\cite{Perazzi2016}.  However this data is considerably sparser than human data, and must be ``shared'' across species, meaning the number of examples for a given animal shape class is considerably fewer than is available for an equivalent variation in human shape.  While segmentation data can be supplied by non-specialist human labellers, it is more difficult to obtain {\em joint position} data.  Some joints are easy to label, such as ``tip of snout'', but others such as the analogue of ``right elbow'' require training of the operator to correctly identify across species.

Of more concern however, is 3D skeleton data.  For humans, motion capture (mocap) can be used to obtain long sequences of skeleton parameters (joint positions and angles) from a wide variety of motions and activities.
For animal tracking, this is considerably harder: animals behave differently on treadmills than in their quotidian environments, and although some animals such as horses and dogs have been coaxed into motion capture studios~\cite{wilhelm2015furyexplorer}, it remains impractical to consider mocap for a family of tigers at play.

These concerns are of course alleviated if we have access to synthetic training data.  Here, humans and animals share an advantage in the availability of parameterized 3D models of shape and pose.  The recent publication of the Skinned Multi-Animal Linear (SMAL) model~\cite{zuffi2017menagerie} can generate a wide range of quadruped species, although without surface texture maps.  However, as with humans, it remains difficult to generate RGB images which are sufficiently realistic to train modern machine learning models.  In the case of humans, this has been overcome by generating depth maps, but this then requires a depth camera at test time~\cite{shotton-kinect}. The alternative, used in this work, is to generate 2D silhouette images so that machine learning will predict joint heatmaps from silhouettes only.

Taking into account the above constraints, this work applies a novel strategy to animal tracking, which assumes a machine-learning approach to extraction of animal silhouettes from video, and then fits a parameterized 3D model to silhouette sequences.  We make the following contributions:
\begin{itemize}
\item A machine-learned mapping from silhouette data of a large class of quadru\-peds to generic 2D joint positions.
\item A novel optimal joint assigment (OJA) algorithm extending the bipartite matching of Cao {\em et al.}~\cite{cao2017realtime} in two ways, one which can be cast as a quadratic program (QP), and an extension optimized using a genetic algorithm (GA).
\item A procedure for optimization of a 3D deformable model to fit 2D silhouette data and 2D joint positions, while encouraging temporally coherent outputs.
\item We introduce a new benchmark animal dataset of joint annotations (BADJA) which contains sparse keypoint labels and silhouette segmentations for eleven animal video sequences. 
Previous work in 3D animal reconstruction has relied on bespoke hand-clicked keypoints~\cite{zuffi2017menagerie,zuffi_lions} and little quantitative evaluation of performance could be given.
The sequences exhibit a range of animals, are selected to capture a variety of animal movement and include some challenging visual scenarios such as occlusion and motion blur.
\end{itemize}
The system is outlined in \figref{overview}.  The remainder of the paper describes related literature before a detailed description of system components.  Joint accuracy results at multiple stages of the pipeline are reported on the new BADJA dataset, which contains ground truths for real animal subjects. We also conduct experiments on synthetic animal videos to produce joint accuracy statistics and full 3D mesh comparisons. A qualitative comparison is given to recent work~\cite{zuffi2017menagerie} on the related single-frame 3D shape and pose recovery problem. The paper concludes with an assessment of strengths and limitations of the work.
\newpage
\section{Related work}
3D animal tracking is relatively new to the computer vision literature, but animal breed identification is a well studied problem~\cite{imagenet_cvpr09}. Video tracking benchmarks often use animal sequences~\cite{DAVIS2017-1st,DAVIS2017-2nd}, although the tracking output is typically limited to 2D affine transformations rather than the detailed 3D mesh that we propose.  Although we believe our work is the first to demonstrate dense 3D tracking of animals in video without the need for user-provided keypoints, we do build on related work across computer vision:

\subsubsection*{Morphable shape models.}
Cashman and Fitzgibbon~\cite{cashman2013shape} obtained one of the first 3D morphable animal models, but their work was limited to small classes of objects (e.g. dolphins, pigeons), and did not incorporate a skeleton.  Their work also showed the use of the 2D silhouette for fitting, which is key to our method. 
Reinert {\em et al.} \cite{reinert2016animated} meanwhile construct 3D meshes by fitting generalized cylinders to hand-drawn skeletons.
Combined skeletal and morphable models were used by Khamis {\em et al.}~\cite{hand-shape} for modelling the human hand, and Loper {\em et al.}~\cite{loper2015smpl} in the SMPL model which has been extensively used for human tracking. 

The SMPL model was extended to animals by Zuffi {\em et al.}~\cite{zuffi2017menagerie}, where the lack of motion capture data for animal subjects is cleverly overcome by building the model from $41$ 3D scans of toy figurines from five quadruped families in arbitrary poses. Their paper demonstrates single-frame fits of their model to real-world animal data, showing that despite the model being built from ``artists' impressions'' it remains an accurate model of real animals. This is borne out further by our work.  Their paper did however depend on per-frame human annotated keypoint labels, which would be costly and challenging to obtain for large video sequences. This work was recently extended~\cite{zuffi_lions} with a refinement step that optimizes over model vertex positions. This can be considered independent to the initial SMAL model fit and would be trivial to add to our method.

\subsubsection*{Shape from silhouette.} Silhouette images have been shown to contain sufficient shape information to enable their use in many 3D recovery pipelines. Chen {\em et al.}~\cite{chen2010inferring} demonstrate single-view shape reconstruction from such input for general object classes, by building a shape space model from 3D samples. More related to our work, Favreau {\em et al.}~\cite{favreau2004animal} apply PCA to silhouette images to extract animal gaits from video sequences. The task of predicting silhouette images from 2D input has been effectively used as a proxy for regressing 3D model parameters for humans~\cite{indirect2017,hmrKanazawa17} and other 3D objects~\cite{wiles2017silnet}.

\subsubsection*{Joint position prediction.} There is an extensive body of prior work related to joint position prediction for human subjects. Earlier work used graphical approaches such as pictorial structure models~\cite{andriluka2010monocular,pishchulin2013poselet,johnson2010clustered}, which have since been replaced with deep learning-based methods~\cite{cao2017realtime,bulat2016human}. Few works predict animal joint positions directly owing to the lack of annotated data, although Mathis {\em et al.}~\cite{mathis2018deeplabcut} demonstrate the effectiveness of human pose estimation architectures for restricted animal domains. Our method instead trains on silhouette input, allowing the use of synthetic training imagery. The related task of animal part segmentation~\cite{wang2015joint,wang2015semantic} has seen some progress due to general object part datasets~\cite{chen_cvpr14,zhou2017scene}.

\subsection{Preliminaries}
\def\R#1{{\mathbb{R}^{#1}}}
\def\RR#1#2{{\mathbb{R}^{#1 \times #2}}}
\def\posn{\phi}
\def\pose{\theta}
\def\npose{P}
\def\shape{\beta}
\def\nshape{B}
\def\verts{\nu}
\def\nverts{V}
\def\jointselect{\mathtt{K}}
\def\njoints{J}
We are given a deformable 3D model such as SMAL~\cite{zuffi2017menagerie} which parametrizes a 3D mesh as a function of {\em pose} parameters~$\pose \in \R\npose$ (e.g.\ joint angles) and {\em shape} parameters~$\shape \in \R\nshape$. 
In detail, a 3D mesh is an array of vertices $\verts \in \RR 3\nverts$ (the vertices are columns of a $3 \times \nverts$ matrix) and a set of triangles represented as integer triples $(i,j,k)$, which are indices into the vertex array.
A deformable model such as SMPL or SMAL may be viewed as supplying a set of triangles, and a function
\begin{equation}
\verts(\pose, \shape) : \R \npose \times \R \nshape \mapsto \RR 3 \nverts
\end{equation}
which generates the 3D model for a given pose and shape.
The mesh topology (i.e.~the triangle vertex indices) is provided by the deformable model, and is the same for all shapes and poses we consider, so in the sequel we shall consider a mesh to be defined only by the 3D positions of its vertices.

In any given image, the model's 3D {\em position} (i.e.\ translation and orientation) is also unknown, and will be represented by a parametrization $\posn$ which may be for example translation as a 3-vector and rotation as a unit quaternion. Application of such a transformation to a $3\times\nverts$ matrix will be denoted by $*$, so that 
\begin{equation}
\posn * \verts(\pose, \shape)
\end{equation}
represents a 3D model of given pose and shape transformed to its 3D position.

\def\proj{\pi}
We will also have occasion to talk about model {\em joints}.  These appear naturally in models with an explicit skeleton, but more generally they can be defined as some function mapping from the model parameters to an array of 3D points analogous to the vertex transformation above.  We consider the joints to be defined by post-multiplying by a $\nverts \times \njoints$ matrix $\jointselect$.  The $j^{\text{th}}$ column of~$\jointselect$ defines the 3D position of joint~$j$ as a linear combination of the vertices (this is quite general, as $\verts$ may include vertices not mentioned in the triangulation).  A general camera model is described by a function $\proj: \R 3 \mapsto \R 2$.  This function incorporates details of the camera intrinsics such as focal length, which are assumed known.  
Thus 
\begin{equation}
\kappa(\posn, \pose, \shape) := \proj(\posn * \verts(\pose, \shape) \jointselect)
\end{equation}
is the $2\times \njoints$ matrix whose columns are 2D joint locations corresponding to a 3D model specified by (position, pose, shape) parameters $(\posn, \pose, \shape)$.

The model is also assumed to be supplied with a rendering function $R$ which takes a vertex array in camera coordinates, and generates a 2D binary image of the model silhouette.  That is,
\begin{equation}
R\bigl(\posn * \verts(\pose, \shape)\bigr) \in \mathbb{B}^{W\times H}
\end{equation}
for an image resolution of $W \times H$.  We use the differentiable renderer of Loper {\em et al.}~\cite{loper2014opendr} to allow derivatives to be propagated through $R$.  
\newpage
\section{Method}
\def\seq#1#2#3#4{\left[{#1_{#2}}\right]_{#2=#3}^{#4}}
The test-time problem to be solved is to take a sequence of input images
$
\mathcal I = \seq I t1T
$
which are segmented to the silhouette of a single animal (i.e.~a video with multiple animals is segmented multiple times), producing a sequence of binary silhouette images 
$
\mathcal S = \seq S t1T.
$

The computational task is to output for each image the shape,  pose, and position parameters describing the animal's motion.

As outlined above, the method has three parts.  
(1.) The discriminative front-end extracts silhouettes from video, and then uses the silhouettes to predict 2D joint positions, with multiple candidates per joint. 
(2.) Optimal joint assignment (OJA) corrects confused or missing skeletal predictions by finding an optimal assignment of joints from a set of network-predicted proposals. Finally, (3.) a generative deformable 3D model is fitted to the silhouettes and joint candidates as an energy minimization process.

\subsection{Prediction of 2D joint locations using multimodal heatmaps}
The goal of the first stage is to take, for each video frame, a $W\times H$ binary image representing the segmented animal, and to output a $W \times H \times \njoints$ tensor of heatmaps. The network architecture is standard: a stacked hourglass network~\cite{newell2016stacked} using synthetically generated training data, but the training procedure is augmented using ``multi-modal'' heatmaps. 

For standard unimodal heatmaps, training data comprises $(S, \kappa)$ pairs, that is pairs of binary silhouette images, and the corresponding 2D joint locations as a $2\times J$ matrix.  To generate each image, a random shape vector $\shape$, pose parameters $\pose$ and camera position $\posn$ are drawn, and used to render a silhouette $R\bigl(\posn * \verts(\pose, \shape)\bigr)$ and 2D joint locations $\kappa(\posn,\pose,\shape)$, which are encoded into a $W \times H \times \njoints$ tensor of heatmaps, blurring with a Gaussian kernel of radius $\sigma$. 

The random camera positions are generated as follows: the orientation of the camera relative to the animal is uniform in the range $[0, 2\pi]$, the distance from the animal is uniform in the range 1 to 20 meters and the camera height is in the range $[0,\frac{\pi}{2}]$. This smaller range is chosen to restrict unusual camera elevation. Finally, the camera ``look" vector is towards a point uniformly in a 1m cube around the animal's center, and the ``up" vector is Gaussian around the model Y axis.  

This training process generalizes well from synthetic to real images due to the use of the silhouette, but the lack of interior contours in silhouette input data often results in confusion between joint ``aliases'': left and right or front and back legs.  When these predictions are wrong and of high confidence, little probability mass is assigned to the area around the correct leg, meaning no available proposal is present after non-maximal suppression.

\begin{figure}[t]
\begin{floatrow}
\ffigbox{
\centering
\begin{tabular}{c}
\includegraphics[trim={4cm 10cm 4cm 10cm},clip,width=0.6\linewidth]{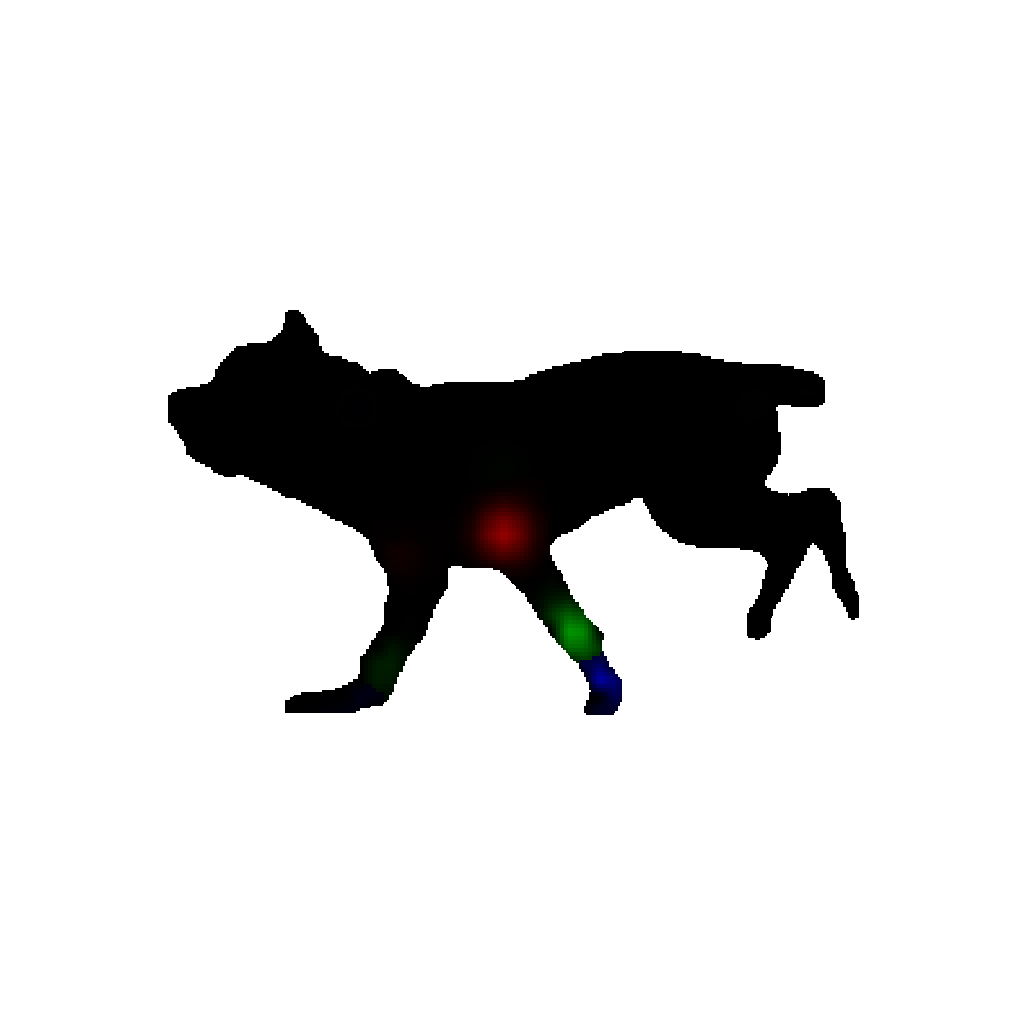}\\
\includegraphics[trim={4cm 10cm 4cm 10cm},clip,width=0.6\linewidth]{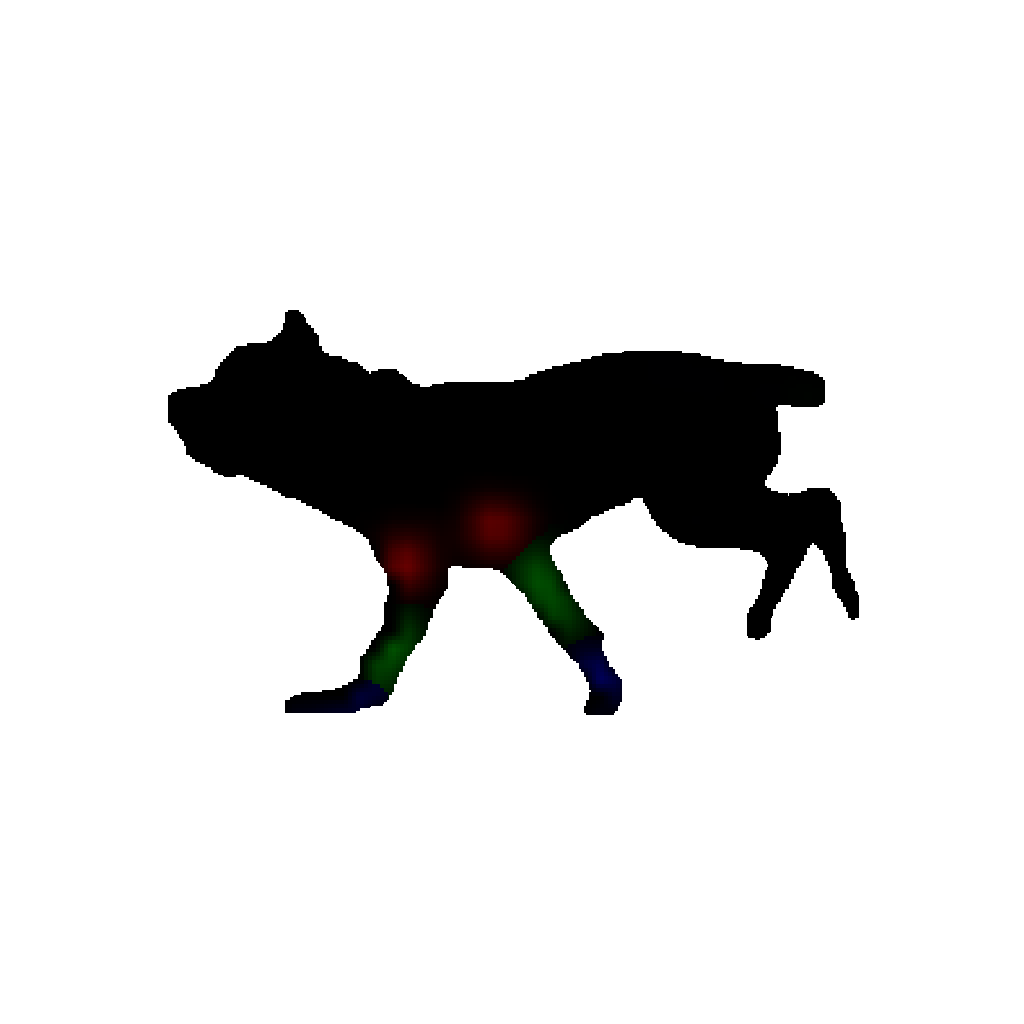}
\end{tabular}
}{
\caption{Example predictions from a network trained on unimodal (top) and multi-modal (bottom) ground-truth  for front-left leg joints.}
\label{fig:single_multi}
}
\ffigbox{
\centering
\def\lp#1#2{\labelledpic{\includegraphics[height=19mm]{#2}}{#1}}
\begin{tabular}{cc}
\lp a{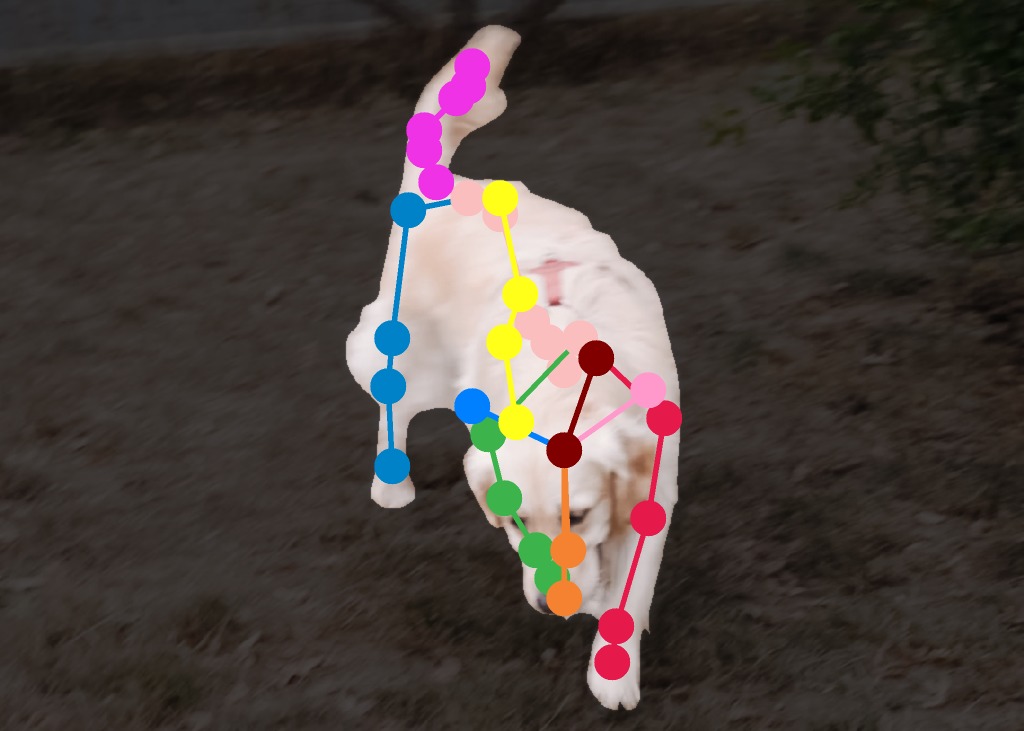}&
\lp b{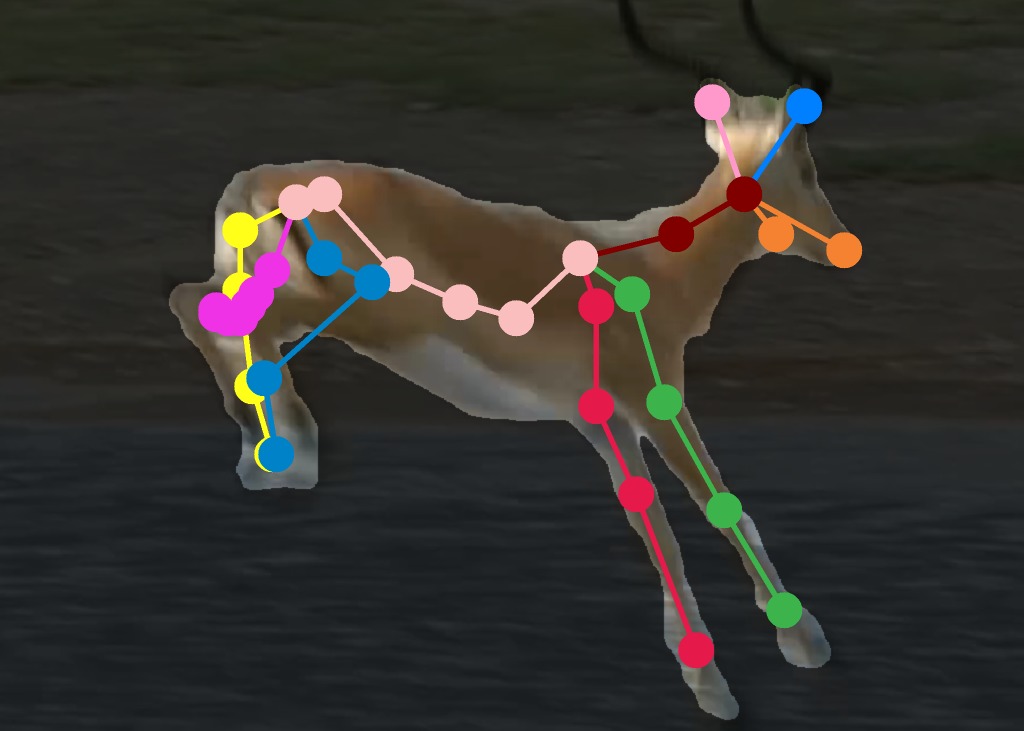}\\
\lp c{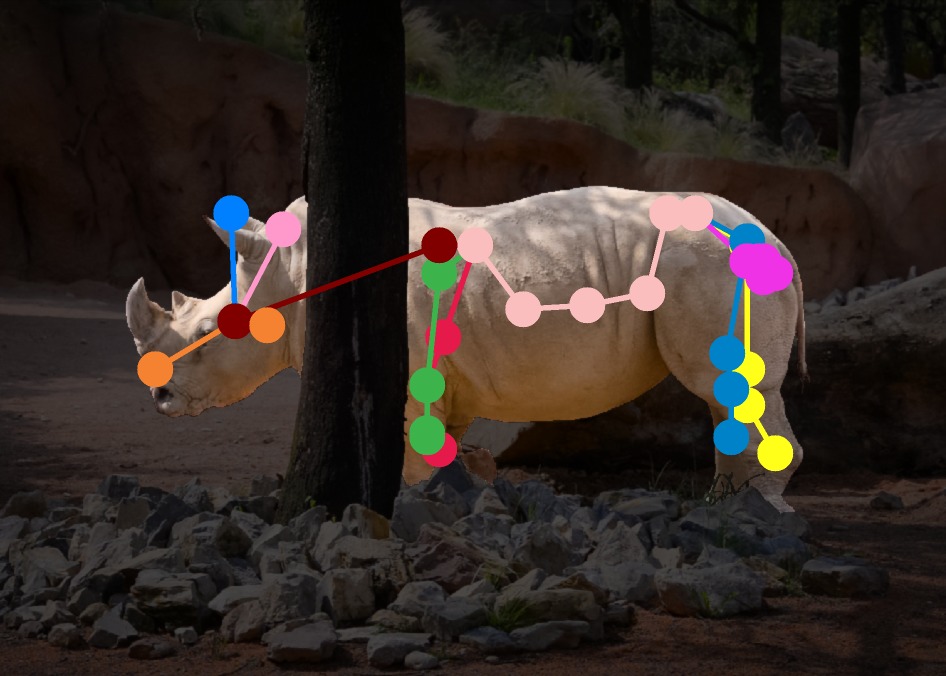}&
\lp d{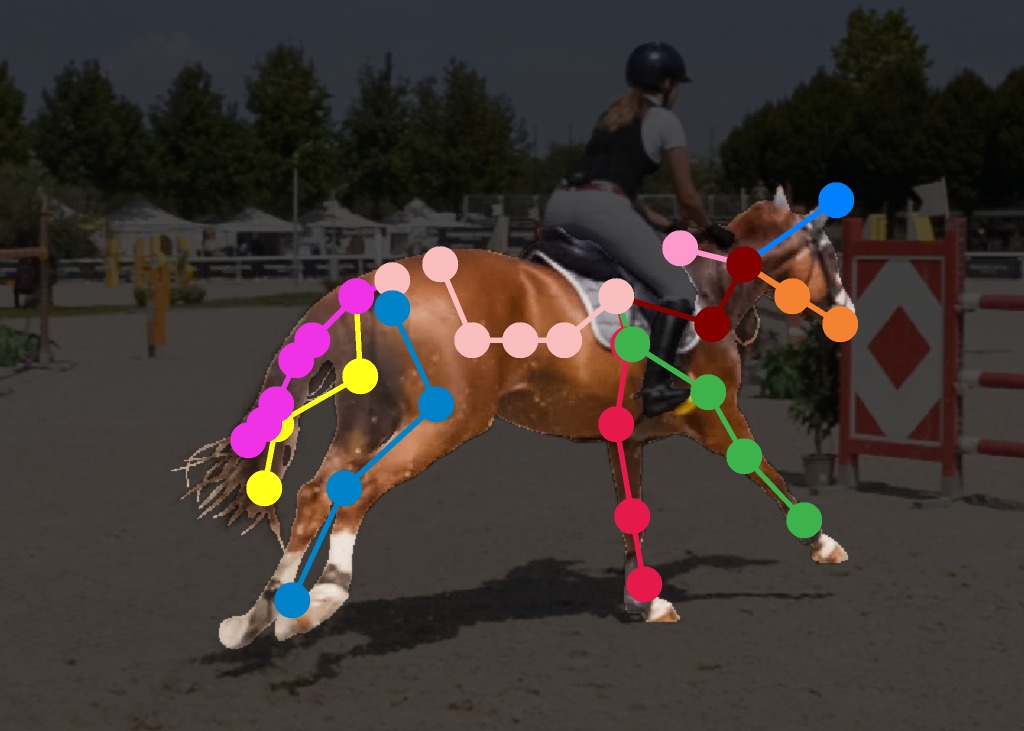}
\end{tabular}
}{
\caption{Example outputs from the joint prediction network, with maximum likelihood predictions linked into skeleton.}
\label{fig:exp-network}
}
\end{floatrow}
\end{figure}

We overcome this by explicitly training the network to assign some probability mass to the ``aliased'' joints. For each joint, we define a list of potential aliases as weights $\lambda_{j,j'}$ and linearly blend the unimodal heatmaps $G$ to give the final training heatmap $H$:
\begin{equation}
    H_{j}(p) = \sum_{j'} \lambda_{j,j'} G(p; \kappa_{j'}, \sigma)
\end{equation}
For non-aliased joints $j$ (all but the legs), we simply set $\lambda_{j,j} = 1$ and $\lambda_{j,j'} = 0$, yielding the unimodal maps, and for legs, we use 0.75 for the joint, and 0.25 for the alias.  We found this ratio sufficient to ensure opposite legs have enough probability mass to pass through a modest non-maximal suppression threshold without overly biasing the skeleton with maximal predicted confidence. An example of a heatmap predicted by a network trained on multimodal training samples is illustrated in \figref{single_multi}. 

\subsection{Optimal joint assignment (OJA)}
\label{sec:qp}
Since heatmaps generated by the joint predictor are multi-modal, the non-maximum suppression procedure yields multiple possible locations for each joint. We represent the set of joint proposals $X = \{x_{jp}\}$, where $x_{jp}$ indicates the 2D position of proposal $p \in \{1,...,N_j\}$ associated with joint $j \in J$.
Before applying the optimizer, we must select a subset of proposals $X^* \subseteq X$ which form a complete skeleton, i.e. precisely one proposal is selected for every joint. In this section we consider how to choose the optimal subset by formulating the problem as an extended optimal assignment problem.

\def\bvec#1{\bar{#1}}
\def\LL#1{L_{\text{#1}}}
In order to select a complete skeleton proposal from the set of joint proposals $\{x_{jp}\}$, we introduce a binary indicator vector $\bvec a_j = \{a_{jp}\} \in \{0, 1\}^{N_j+1}$, where $a_{jp} = 1$ indicates that the $p^\text{th}$ proposal for joint $j$ is a correct assignment, and the $p = N_j+1$ position corresponds to a {\em null proposal}, indicating that joint $j$ has no match in this image.
The null proposals are handled as described in each of the energy terms below.
Let $A$ be the jagged array $[\bvec a_j]_{j=1}^J$ containing all assignment variables (for the current frame), and let $X^* = X(A)$ denote the subset of points selected by the binary array $A$.   
Optimal assignment minimizes the function
\begin{equation}
L(A) = \LL{prior}(A) + \LL{conf}(A) + \LL{temp}(A) + \LL{cov-sil}(A) + \LL{cov-bone}(A)
\end{equation}
which balances agreement of the joint configuration with a learned {\em prior}, the network-supplied {\em confidences}, {\em temporal} coherence, and {\em coverage} terms which encourage the  model to correctly project over the silhouette.   Without the coverage terms, this can be optimized as a quadratic program, but we obtain better results by using the coverage terms, and using a genetic algorithm.  In addition, the parameters $A$ must satisfy the $J$ constraints $\sum_{p=1}^{N_j+1} a_{jp} = 1$, that exactly one joint proposal (or the null proposal) must be selected for each joint.

\def\lterm#1{\subsubsection{$\LL{#1}$:}}

\lterm{prior} We begin by defining the prior probability of a particular skeletal configuration as a multivariate Gaussian distribution over selected joint positions.

The mean $\mu \in \R{2J}$ and covariance $\Sigma \in \R{2J\times 2J}$ terms are obtained from the training examples generated as above. The objective of OJA is to select a configuration $X^*$ which maximizes the prior, which is equivalent to minimizing the Mahalanobis distance
$(x^*-\mu)^T\Sigma^{-1}(x^*-\mu)$, which is given by the summation
\begin{equation}
\LL{prior}(A) = \sum_j^J\sum_p^{N_j}\sum_k^J\sum_q^{N_k}a_{jp}a_{kq}(x_{jp} - \mu_j)\Sigma_{jk}^{-1}(x_{kq}-\mu_k)
\end{equation}
This is a quadratic function of $A$, so $\LL{prior}(A) = \text{vec}(A)^\top Q \text{vec}(A)$ for a fixed matrix $Q$, and can be formulated as a quadratic program (QP).  Null proposals are simply excluded from the sum, equivalent to marginalizing over their position. 

\lterm{conf}
The next energy term comes from the output of the joint prediction network, which provides a confidence score $y_{jp}$ associated with each joint proposal~$x_{jp}$.  Then $\LL{conf}(A) = \sum_j\sum_p -\lambda\log(y_{jp}) a_{jp}$ is a linear function of $A$, 
and $\lambda_{\text{conf}}$ is a tunable parameter to control the relative contribution of the network confidences compared with that of the skeleton prior.
Null proposals pay a fixed cost $\lambda_{null}$, effectively acting as a threshold whereby the null proposal will be selected if no other proposal is of sufficient likelihood. 

\lterm{temp}
A common failure case of the joint prediction network is in situations where a joint position is highly ambiguous, for example between the left and right legs. In such cases, the algorithm will commonly alternate between two equally likely predictions. This leads to large displacements in joint positions between consecutive frames which are difficult for the later model fitting stage to recover from. This can be addressed by introducing a temporal term into the OJA. We impose a prior on the distance moved by each joint between frame $t_0$ and $t_1$, which is given by a normal distribution with zero mean and variance $\sigma^{2} =e^{\tau|t_1 - t_0 - 1|}$. 
The parameter $\tau$ controls the strength of the interaction between distant frames. This results in an additional quadratic term in our objective function, which has the form $L_{temp} = a^\top T^{(t_0, t_1)} a$ for matrix $T^{(t_0, t_1)}$ given by 
\begin{equation}
\left[T^{(t_0, t_1)}\right]_{jp, kq} = \begin{cases}
e^{-\alpha|t_1 - t_0 - 1|}||x^{(t_0)}_{jp} - x^{(t_1)}_{kq}||^2 & \text{if } j=k\\
0 & \text{otherwise}
\end{cases}
\end{equation}

\subsubsection*{QP solution.}
Thus far, all terms in $L(A)$ are quadratic or linear.
To optimize over an entire sequence of frames, we construct the block diagonal matrix $\hat{Q}$ whose diagonal elements are the prior matrices $Q^{(t)}$ and the block symmetric matrix $\hat{T}$ whose off-diagonal elements are the temporal matrices $T^{(t_0, t_1)}$. The solution vector for the sequence $\hat{A}$ is constructed by stacking the corresponding vectors for each frame. The quadratic program is specified using the open source CVXPY library \cite{diamond2016cvxpy} and solved using the ``\emph{Suggest-and-Improve}'' framework proposed by Park and Boyd \cite{park2017general}. It is initialized by choosing the proposal with the highest confidence for each joint. Appropriate values for the free parameters $\lambda_{\text{conf}, \text{temp}}$ and $\alpha$ were chosen empirically via grid search. 

\lterm{cov-\{sil,bone\}}
The above quadratic formulation is sufficient to correct many errors in the raw output (which we later demonstrate in the experimental section), but suffers from an `overcounting' problem, in which leg joint predictions both cover the same silhouette leg region, leaving another leg empty. We therefore extend the definition of $L(A)$ to include two additional terms. 

\def\silhouette{S}

\lterm{cov-sil} penalizes large silhouette areas with no nearby selected joint. This term requires a precomputed set of silhouette sample points $Z \subseteq \mathbb{R}^2$, which we aim to ``cover'' as best as possible with the set of selected joints. Intuitively, the silhouette is considered well-covered if all sample points are close to \emph{some} selected joint proposal. The set $Z$ is generated from the medial axis transform (MAT)\cite{blum1967transformation} of the silhouette, $Z^{t} = \text{MAT}(\silhouette^{t})$
with a cubed loss strongly penalizing projection outside the silhouette:
\begin{equation}
\LL{cov-sil}(A^{t};X^{t},Z^{t}) = \sum_{i}\min_{j}\|Z_{i}^{t} - \hat{X}_{j}^{t}\|^3
\end{equation}

\begin{figure}[t!]
\begin{floatrow}
\ffigbox{
\def\bb{\rule{2in}{0pt}\rule{0pt}{1in}}
\def\bjb{\rule{0.5in}{0pt}\rule{0pt}{0.25in}}

\begin{center}
\scalebox{-1}[1]{\includegraphics[width=0.49\linewidth]{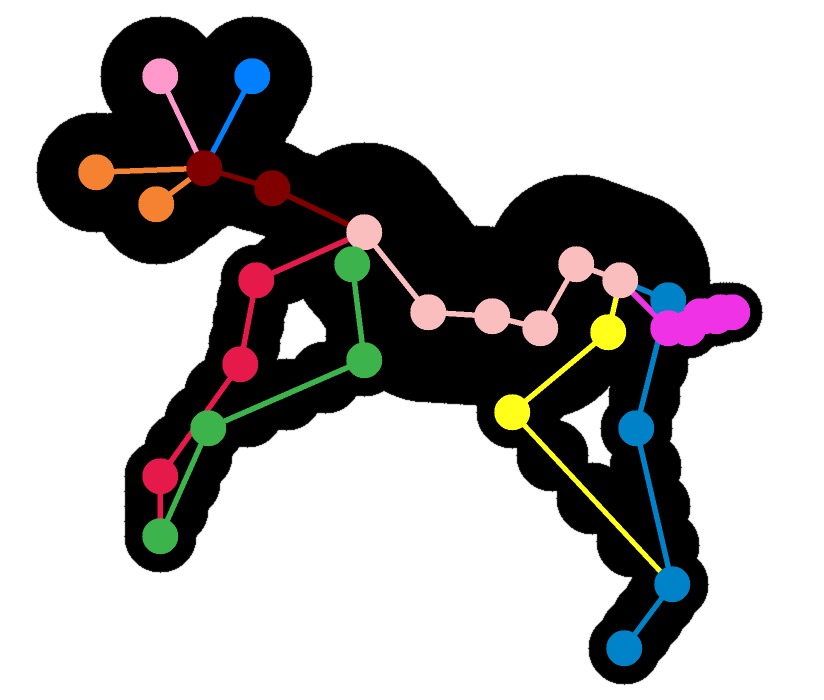}}
\scalebox{-1}[1]{\includegraphics[width=0.49\linewidth]{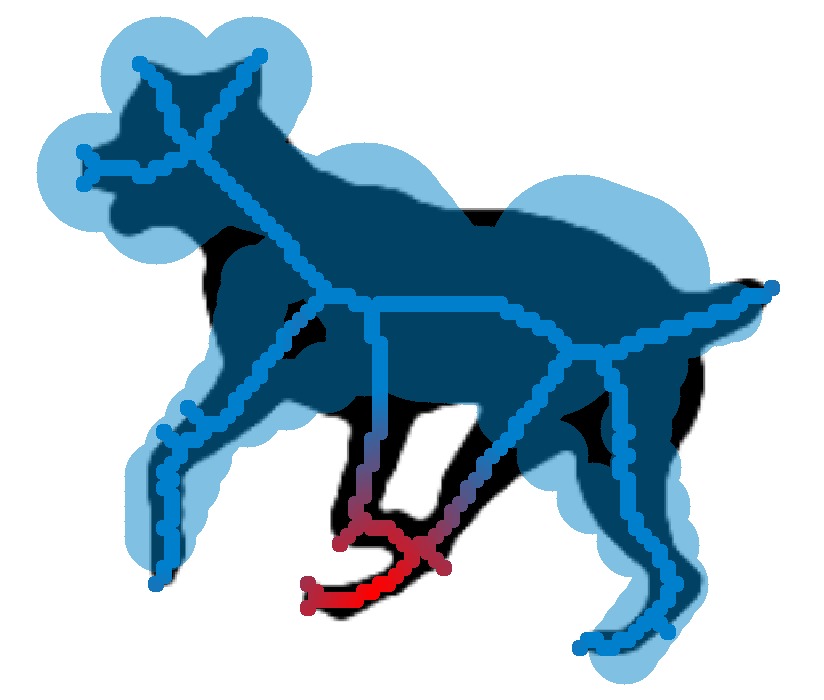}}
\end{center}
}
{\caption{Silhouette coverage loss. The error (shown in red) is the the distance between the median axis transform (right) and the nearest point on an approximate rendering (left).}
\label{fig:example_errors}}
\ffigbox{ 
    \raisebox{1 em}{
    \centering
    \includegraphics[trim={0cm 0cm 0cm 0cm}, clip,width=0.45\linewidth]{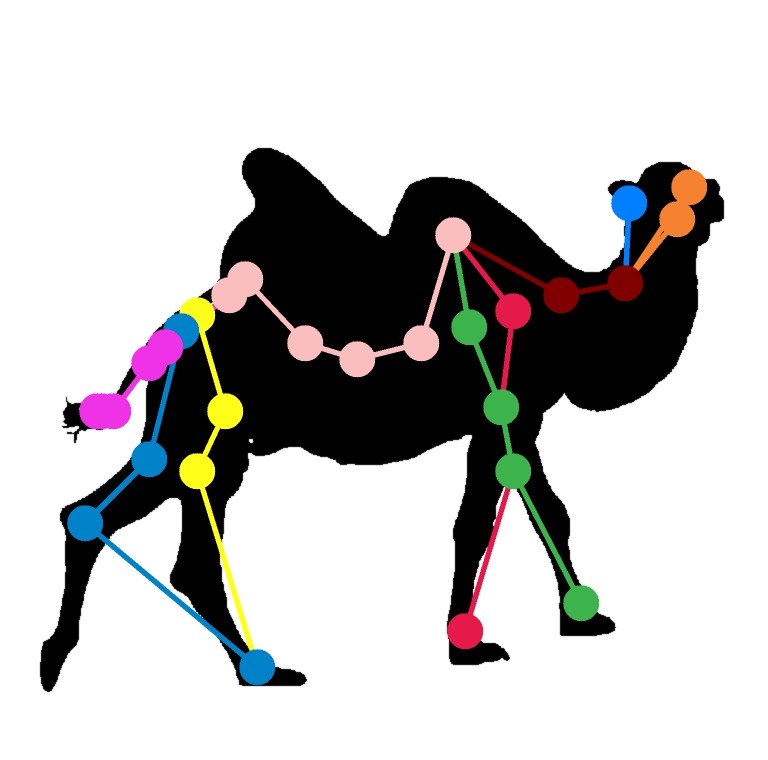}
    \includegraphics[trim={0cm 0cm 0cm 0cm},clip,width=0.45\linewidth]{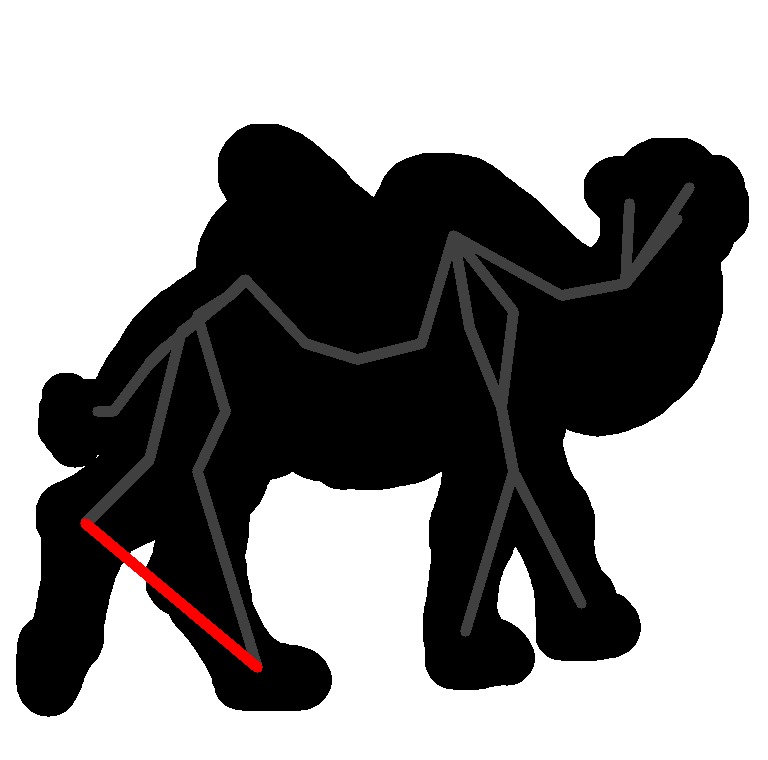}
    }
}
{\caption{Bone coverage loss. One of the back-right leg joints is incorrectly assigned (left), leading to a large penalty since the lower leg bone crosses outside the dilated silhouette (right).}
\label{fig:cov-bone}
}
\end{floatrow}
\end{figure}

\lterm{cov-bone} is used to prevent bones crossing the background. The joint hierarchy is stored in a kinematic tree structure $K = \{\{j,k\} \text{ if joints } j, k \text{ are connected by a bone}\}$.
\begin{equation}
\LL{cov-bone}(A^{t};X^{t},\silhouette^{t},K) = \sum_{\{j,k\} \in K}\biggl(1 - \min_{\lambda \in \big[0:0.1:1\big]}\silhouette^{t}(\hat{X}_{j}^{t} + \lambda(\hat{X}_{j}^{t} - \hat{X}_{k}^{t}))\biggr)
\end{equation}

\begin{figure}[t!]
   
\end{figure}

\subsubsection*{GA Solution.}
We minimize this more complex objective using a genetic algorithm (GA)\cite{holland1992adaptation}, which requires defining a fitness function, ``genes", an initial population, crossover procedure, and mutation procedure. 
The {\em fitness function} is precisely the energy $L(A)$ given above, and the {\em genes} are vectors of $J$ integers, rather than one-hot encodings.
We begin with a population size of 128 genes, in which the first 32 are set equal to the max confidence solutions given by the network in order to speed up convergence. The remaining 96 are generated by selecting a random proposal for each joint. {\em Crossover} is conducted as standard by slicing genes in two parts, and pairing first and second parts from different parents to yield the next generation. In each generation, each gene has some probability of undergoing a {\em mutation}, in which between 1 and 4 joints have new proposals randomly assigned. Weights were set empirically and we run for 1000 generations.
Examples of errors corrected by these two energy terms are shown in Fig.~\ref{fig:example_errors} and Fig.~\ref{fig:cov-bone}.

\subsection{Generative model optimization}
\def\E#1{{E_{\text{#1}}}}
The generative model optimization stage refines model parameters to better match the silhouette sequence $\mathcal S$, by minimizing an energy which sums 4 terms:

\def\ss#1{\vspace{-0ex}\subsubsection{#1}}

\ss{Silhouette energy.}
The silhouette energy $\E{sil}$ compares the rendered model to a given silhouette image, given simply by the L2 difference between the OpenDR rendered image and the given silhouette:
\begin{equation}
\E{sil}(\posn, \pose, \shape; S) = \lVert S - R\bigl(\posn * \verts(\pose, \shape)\bigr) \rVert
\end{equation}

\ss{Prior energy.}
The prior term $\E{prior}$ encourages the optimization to remain near realistic shapes and poses.  It can constrain pose only weakly because of the lack of training data from real poses.
We adopt three prior energy terms from the SMAL model. The Mahalanobis distance is used to encourage the model to remain close to: (1) a distribution over shape coefficients given by the mean and covariance of SMAL training samples of the relevant animal family, (2) a distribution of pose parameters built over a walking sequence. The final term ensures the pose parameters remain within set limits.
\begin{equation}
\E{lim}(\pose) = \max\{\pose - \pose_{\text{max}}, 0\} + \max\{\pose_{\text{min}} - \pose, 0\}.
\end{equation}

\ss{Joints energy.}
The joints energy $\E{joints}$ compares the rendered model joints to the OJA predictions, and therefore must account for missing and incorrect joints.  It is used primarily to stabilize the nonlinear optimization in the initial iterations, and its importance is scaled down as the silhouette term begins to enter its convergence basin.

\begin{equation}
\E{joints}(\posn, \pose, \shape; X^{*}) = 
\lVert X^{*} - \posn * \verts(\pose,\shape)\jointselect(:,j) \rVert
\end{equation}

\ss{Temporal energy.}
The optimizer for each frame is initialized to the result of that previous. In addition, a simple temporal smoothness term is introduced to penalize large inter-frame variation:
\begin{equation}
\E{temp}(\posn, \pose, \shape; X^{*}) = (\phi_t - \phi_{t+1})^2 + (\shape_t - \shape_{t+1})^2
\end{equation}
The optimization is via a second order dogleg method~\cite{lourakis2005levenberg}.

\section{Experiments}
\subsubsection*{Datasets.}
\begin{figure}[t]
\def\bb{\rule{2in}{0pt}\rule{0pt}{1in}}
\begin{center}
\resizebox{.95\linewidth}{!}{
\includegraphics[width=.24\linewidth]{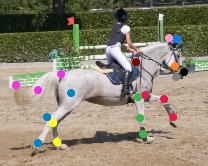}
~
\includegraphics[width=.24\linewidth]{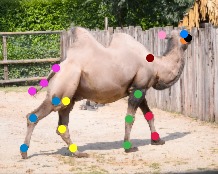}
~
\includegraphics[width=.24\linewidth]{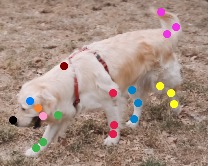}
~
\includegraphics[width=.24\linewidth]{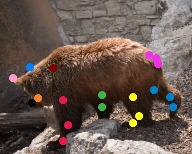}}
\end{center}
\caption{Example joint annotations from the BADJA dataset.  A total of 11 video sequences are in the dataset, annotated every 5 frames with 20 joint positions and visibility indicators.
}
\label{fig:badja_examples}
\end{figure}

In order to quantify our experiments, we introduce a new benchmark animal dataset of joint annotations (BADJA) comprising several video sequences with 2D joint labels and segmentation masks.
These sequences were derived from the DAVIS video segmentation dataset~\cite{Perazzi2016}, as well as additional online stock footage for which segmentations were obtained using Adobe's UltraKey tool~\cite{adobe_ultrakey}. A set of twenty joints on the 3D SMAL mesh were labeled, illustrated in \figref{badja_examples}. These joints were chosen on the basis of being informative to the skeleton and being simple for a human annotator to localize. To make manual annotation feasible and to ensure a diverse set of data, annotations are provided for every fifth frame. 

The video sequences were selected to comprise a range of different quadrupeds undergoing various movement typical of their species. Although the dataset is perhaps insufficient in size to train deep neural networks, the variety in animal shape and pose renders it suitable for evaluating quadruped joint prediction methods. 

\subsection{Joint prediction}
\label{sec:exp-network}
For the joint predictor $\rho$ we train a stacked hourglass network \cite{newell2016stacked}. Following state-of-the-art performance on related human 2D pose estimation datasets (\cite{andriluka14cvpr},~\cite{mscoco}), we construct a network consisting of 8 stacks, 256 features and 1 block. As input we provide synthetically-generated silhouette images of size $256\times 256$, which are obtained by randomly sampling shape and pose parameters from the SMAL model. The corresponding training targets are ground truth heatmaps produced by smoothing the 2D projected joint locations with a Gaussian kernel. Since we are working with synthetic data, we are able to generate training samples on the fly, resulting in an effectively infinite training set. A small adaptation was required to prevent the network degenerating to an unfavourable solution on silhouette input: foreground masks were applied to both ground truth silhouette and predicted heatmaps to prevent the network degenerating to an all-zero heatmap, which produces a reasonably good loss and prevents the network training successfully. The network was trained using the RMSProp optimizer for 40k iterations with a batch size of 18 and learning rate of $2.5\times 10^{-4}$. The learning rate was decayed by 5\% every 10k iterations. Training until convergence took 24 hours on a Nvidia Titan X GPU.

Joint accuracy is evaluated with the Probability of Correct Keypoint (PCK) metric defined by Yang and Ramanan~\cite{yang2013articulated}. The PCK is the percentage of predicted keypoints which are within a threshold distance $d$ from the ground truth keypoint location. The threshold distance is given by $d=\alpha\sqrt{|S|}$ where $|S|$ is the area of the silhouette and $\alpha$ is a constant factor which we set to $\alpha=0.2$ for these experiments.

\begin{figure}[h!]
\centering
\def\p#1{\includegraphics[trim={0 1cm 0 1cm},clip,height=0.12\linewidth] {dog_agility_blooper/#1.jpg}}
\p{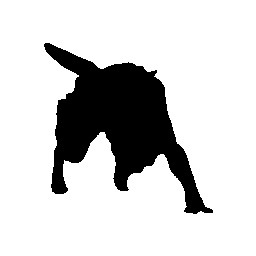}
\p{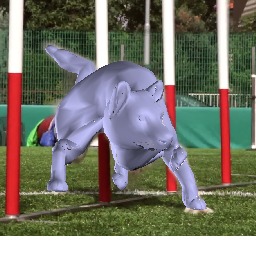}
\def\p#1{\includegraphics[trim={0 1cm 0 1cm},clip,height=0.12\linewidth] {elephant_blooper/#1.jpg}}
\p{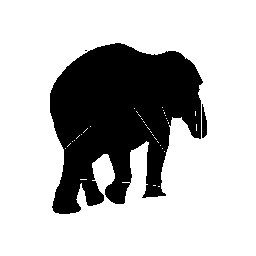}
\p{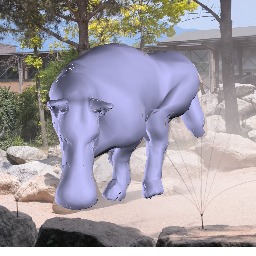}
\def\p#1{\includegraphics[trim={0 1cm 0 1cm},clip,height=0.12\linewidth] {rhino_blooper/#1.jpg}}
\p{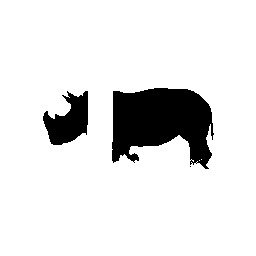}
\p{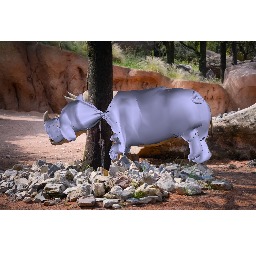}

\caption{Failure modes of the proposed system. \emph{Left}: Missing interior contours prevent the optimizer from identifying which way the dog is facing. \emph{Middle}: The model has never seen an elephant, so assumes the trunk is the tail. \emph{Right}: Heavy occlusion. The model interprets the tree as background and hence the silhouette term tries to minimize coverage over this region.}
\label{fig:blooper}
\end{figure}

\figref{exp-network} shows a selection of maximum likelihood joint predictions on real world images. Note that despite being trained only on synthetic data, the network generalizes extremely well to animals in the wild. The performance extends even to species which were not present in the SMAL model, such as the impala and rhino. The network is also robust to challenging poses (\ref{fig:exp-network}b), occlusions (\ref{fig:exp-network}c) and distraction objects such as the human rider in (\ref{fig:exp-network}d). It is however susceptible to situations where the silhouette image is ambiguous, for example if the animal is facing directly towards or away from the camera. Figure~\ref{fig:blooper} contains examples of failure modes.

\subsection{Optimal joint assignment}
\begin{figure}[t]
\begin{floatrow}
\ffigbox{
\def\bb{\rule{2in}{0pt}\rule{0pt}{1in}}
\def\comparisonheight{20mm}
\begin{tabular}{ccc}
	\includegraphics[trim={7cm 6cm 7cm 6cm},clip,height=\comparisonheight]{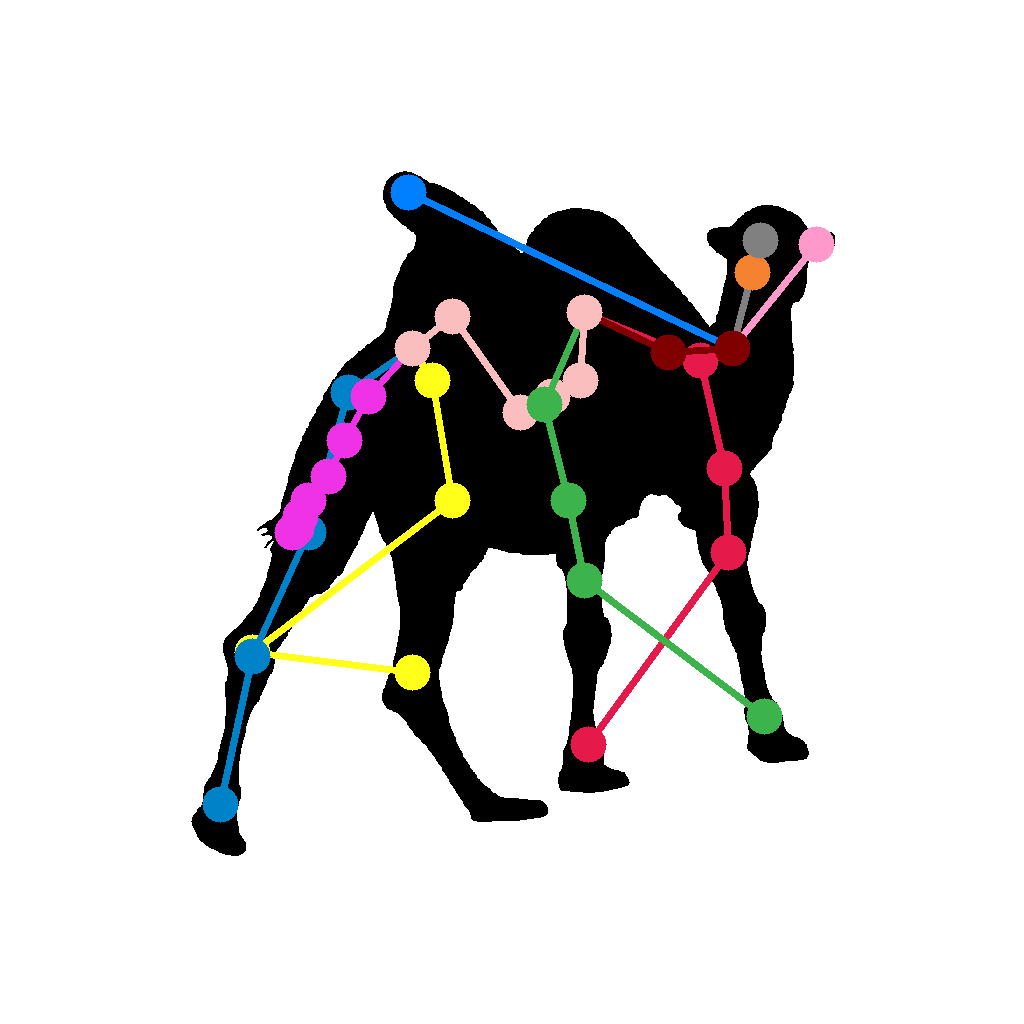} &
	
	\includegraphics[trim={7cm 6cm 7cm 6cm},clip,height=\comparisonheight]{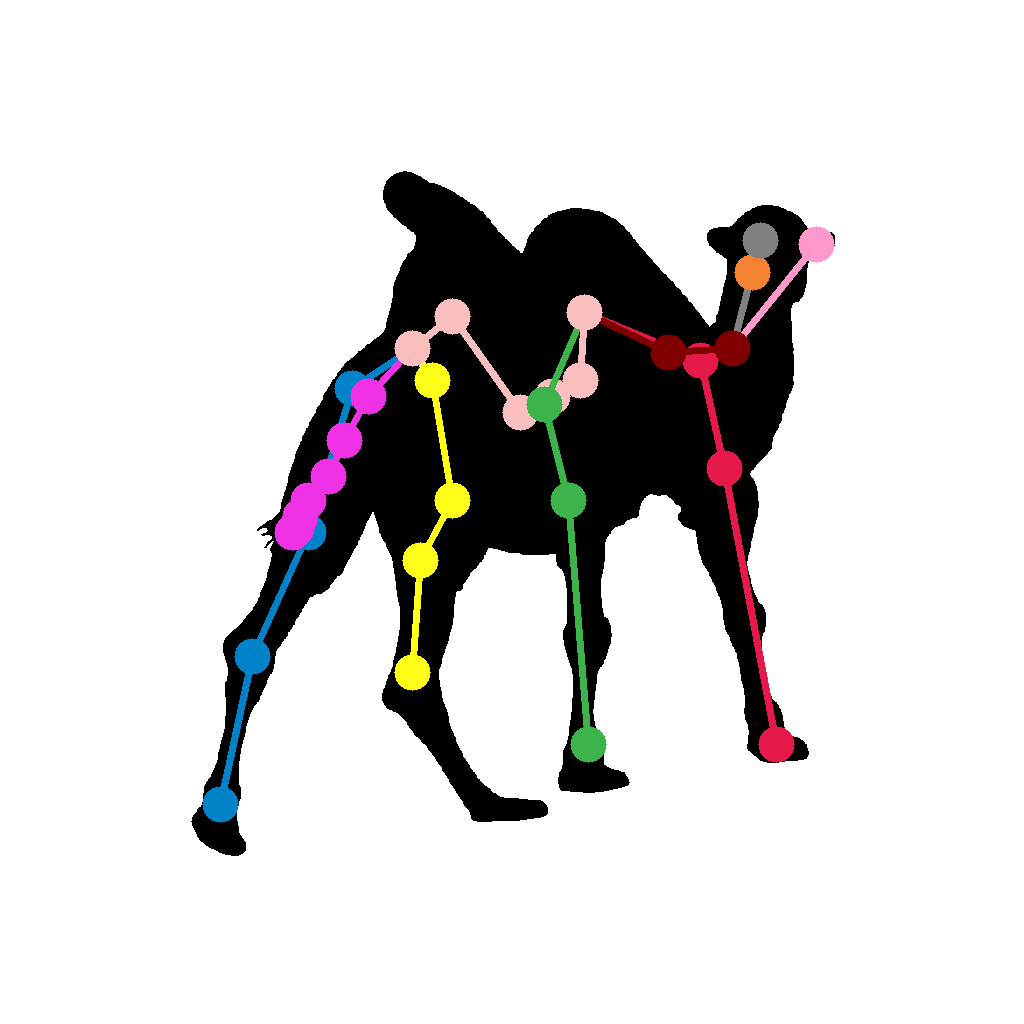} &
	\includegraphics[trim={7cm 6cm 7cm 6cm},clip,height=\comparisonheight]{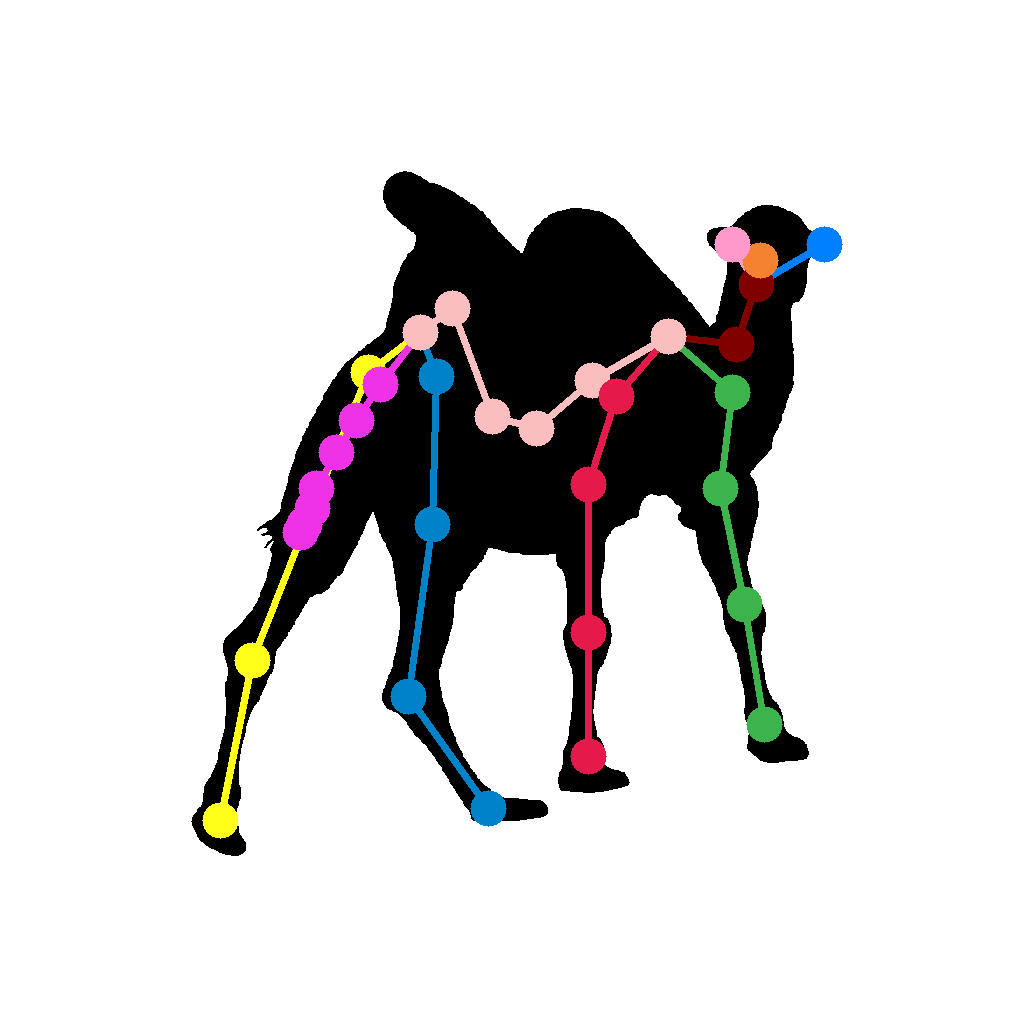} \\
	
    \includegraphics[trim={6cm 6cm 6cm 6cm},clip,height=\comparisonheight]{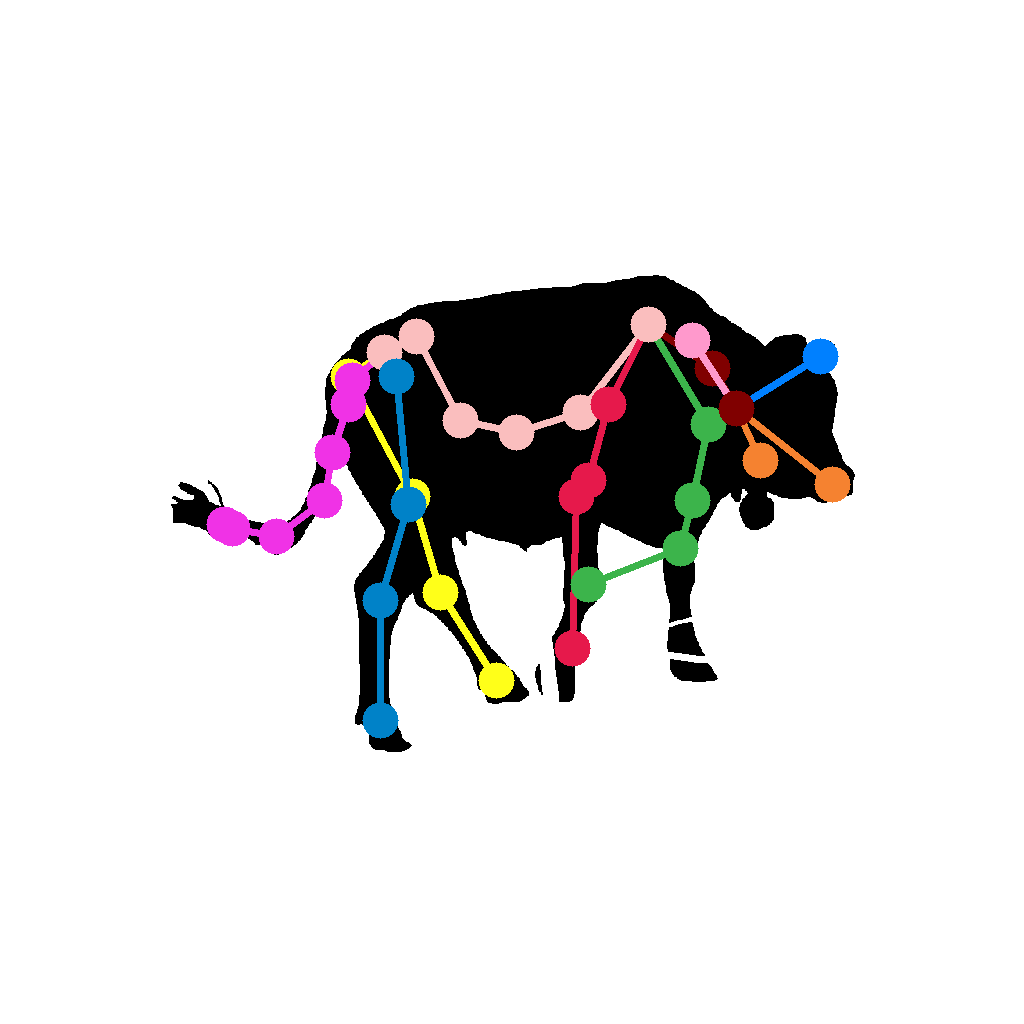}
	 &
    \includegraphics[trim={6cm 6cm 6cm 6cm},clip,height=\comparisonheight]{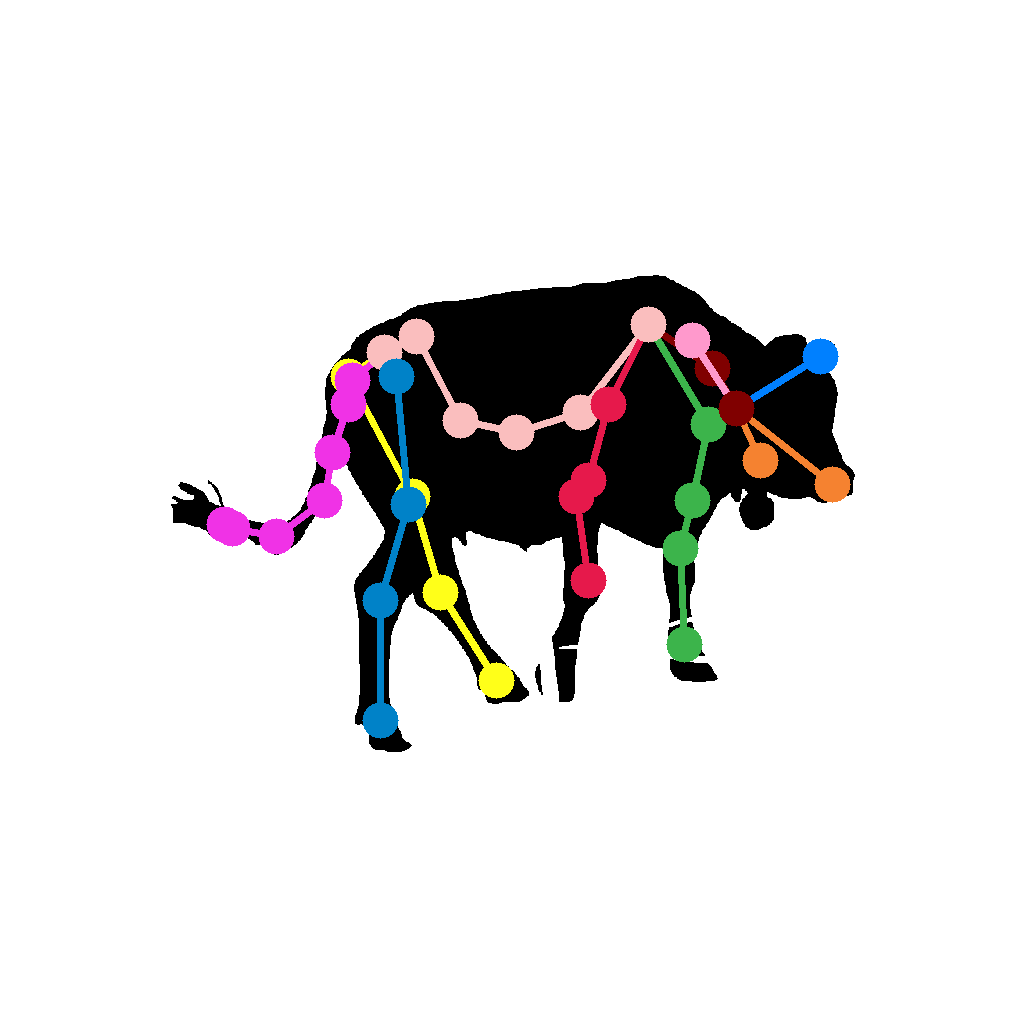}
	 &
    {}\includegraphics[trim={6cm 6cm 6cm 6cm},clip,height=\comparisonheight]{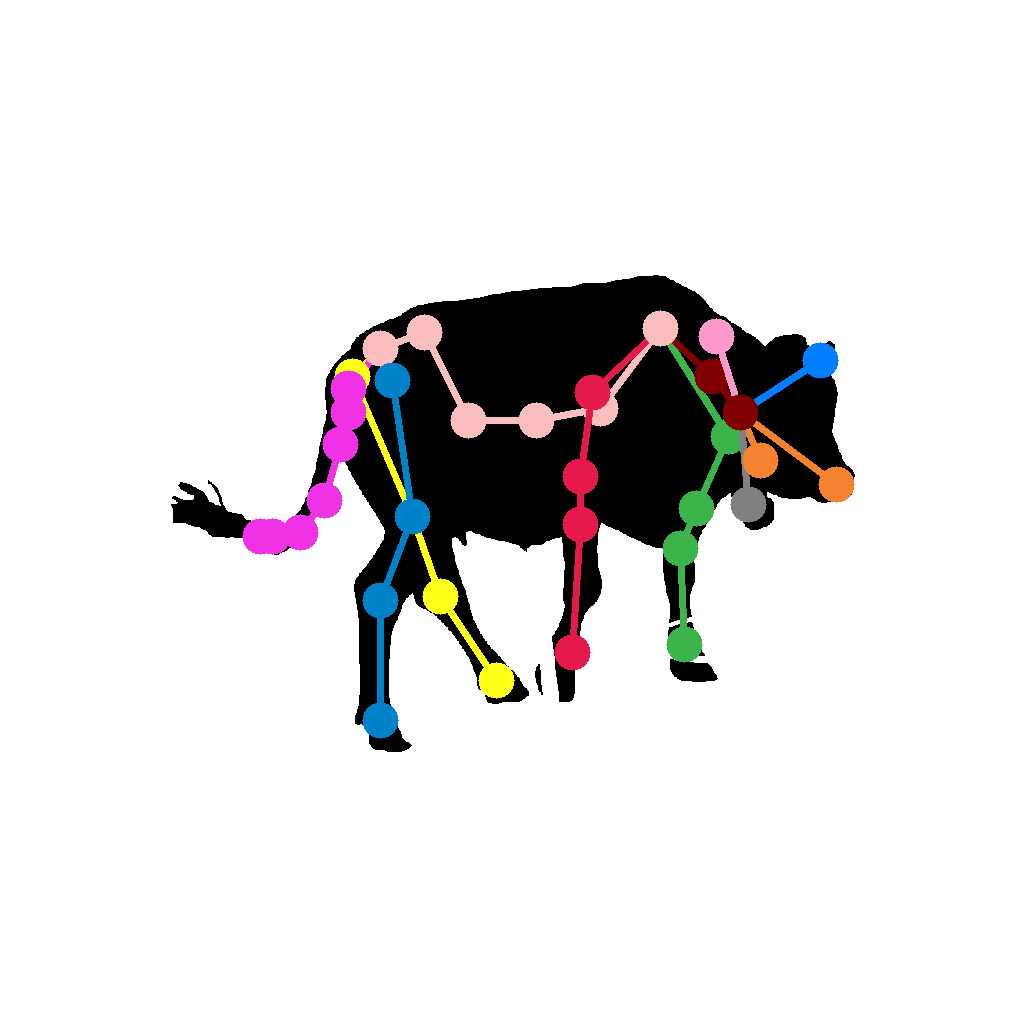} \\
    (a) & (b) & (c)
\end{tabular}
}{
\caption{Example skeletons from raw predictions (a), processed with OJA-QP (b), and OJA-GA (c).}
\label{fig:comparison}
}
~
\capbtabbox{%
\small
\begin{tabular}{lccc}
\toprule
               & Raw       & QP        & GA    \\
\midrule
bear           & 83.1      & 83.7      & \textbf{88.9}   \\
camel          & 73.3      & 74.1      & \textbf{87.1}   \\
cat            & 58.5      & \textbf{60.1}      & 58.4    \\
cows           & 89.2      & 88.4      & \textbf{94.7}  \\
dog            & \textbf{66.9}  & 66.6   & \textbf{66.9}  \\
horsejump-high & 26.5      & \textbf{27.7}      & 24.4   \\
horsejump-low  & 26.9      & 27.0      & \textbf{31.9}   \\
tiger          & 76.5      & 88.8      & \textbf{92.3}   \\
rs\_dog        & 64.2      & 63.4      & \textbf{81.2}   \\
\midrule
Average        & 62.8      & 64.4      & \textbf{69.5}   \\
\bottomrule
\end{tabular}
}{\caption{Accuracy of OJA on BADJA test sequences.}
\label{tab:animal}
}
\end{floatrow}
\end{figure}

Following non-maximum suppression of the joint heatmaps obtained in Section~\ref{sec:exp-network}, we apply OJA to select an optimal set of joints with which to initialize the final optimization stage. It can be seen that the OJA step is able to address many of the failure cases introduced by the joint prediction network, for example by eliminating physically implausible joint configurations (\figref{comparison}, row 1) or by resolving the ambiguity between the left and right legs (\figref{comparison}, row 2).  Table~\ref{tab:animal} summarizes the performance of both the raw network predictions and results of the two OJA methods. Over most of the sequences in the BADJA dataset it can be seen that the use of coverage terms (employed by the OJA-GA model) improves skeleton accuracy. In particular, the bear, camel and rs\_dog sequences show substantial improvements. The method does however struggle on the horsejump\_high sequence, in which part of the silhouette is occluded by the human rider which adversely affects the silhouette coverage term. Across all sequences the selected OJA-GA method improves joint prediction accuracy by 7\% compared to the raw network output. 

\begin{table}[ht]
\centering
\small
\begin{tabular}{@{}cccccccccc@{}}

\toprule
       \multirow{2}{*}{Seq.}  & \multirow{2}{*}{Family} & \multicolumn{2}{c}{PCK (\%)} &   \multirow{2}{*}{Mesh}  & \multirow{2}{*}{Seq.} & \multirow{2}{*}{Family} & \multicolumn{2}{c}{PCK (\%)} &   \multirow{2}{*}{Mesh} \\
&  & Raw                  & OJA-GA &  &&& Raw                  & OJA-GA           \\ \midrule
01 & Felidae         & 91.8     & 91.9  & 38.2   &  06 & Equidae         & 84.4     & 84.8  & 19.2     \\
02 & Felidae         & 94.7     & 95.0  & 42.4   &  07 & Bovidae         & 94.6     & 95.0  & 40.6     \\
03 & Canidae         & 87.7     & 88.0  & 27.3   &  08 & Bovidae        & 85.2     & 85.8  & 41.5     \\  
04 & Canidae         & 87.1     & 87.4  & 22.9   &  09 & Hippopotamidae  & 90.5     & 90.6  & 11.8     \\
05 & Equidae         & 88.9     & 89.8  & 51.6   &  10 & Hippopotamidae  & 93.7     & 93.9  & 23.8     \\    
\bottomrule
\end{tabular}%

\caption{Quantitative evaluation on synthetic test sequences. We evaluate the performance of the raw network outputs and quadratic program post-processing using the probability of correct keypoint (PCK) metric (see sec. \ref{sec:exp-network}). We evaluate mesh fitting accuracy by computing the mean distance between the predicted and ground truth vertices.}
\label{tab:synthetic}
\end{table}

\begin{figure}[t]
\def\bb{\rule{2in}{0pt}\rule{0pt}{1in}}
\def\bjb{\rule{0.5in}{0pt}\rule{0pt}{0.25in}}
\setlength{\fboxsep}{0pt}%
\centering
\begin{tabular}{ccc}
\includegraphics[width=0.3\linewidth]{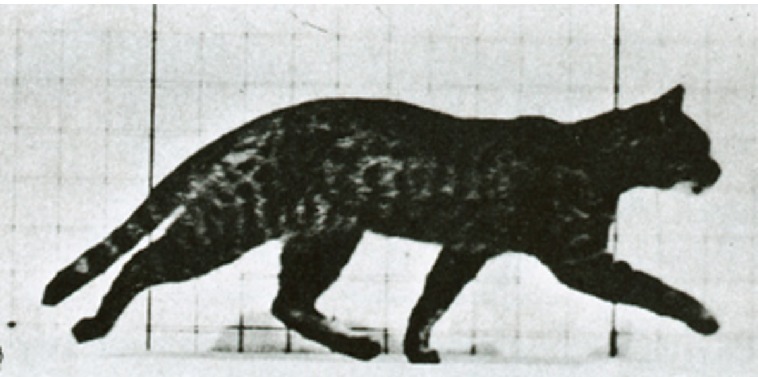}
&
\includegraphics[width=0.3\linewidth]{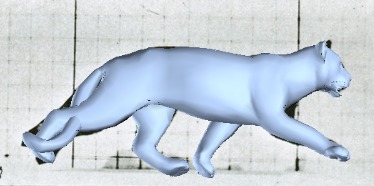}
&
\includegraphics[width=0.3\linewidth]{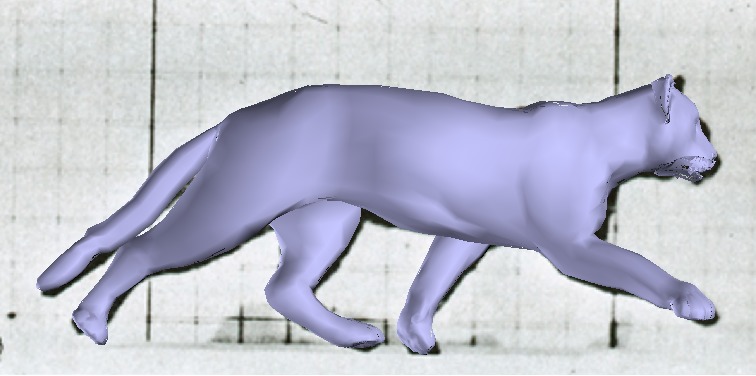}
\\

\includegraphics[width=0.3\linewidth]{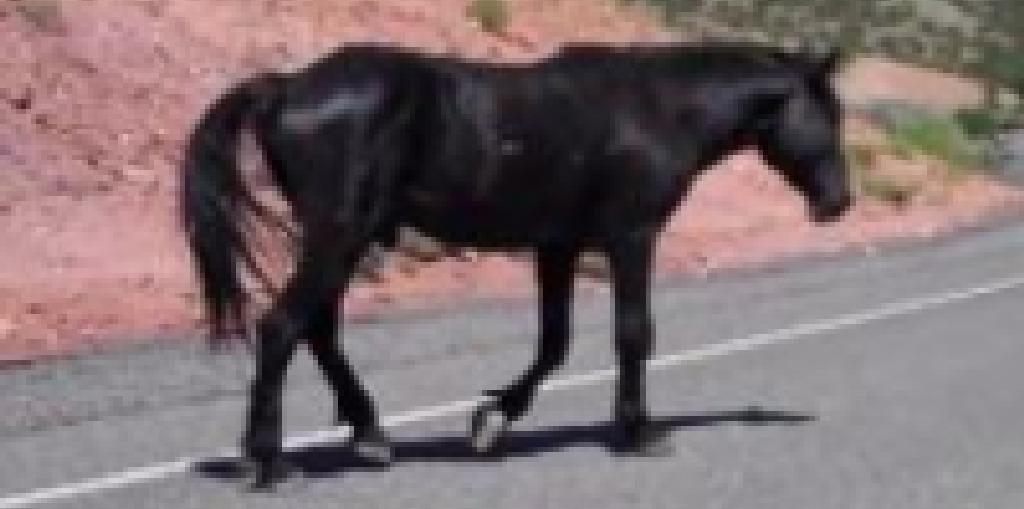} &
\includegraphics[width=0.3\linewidth]{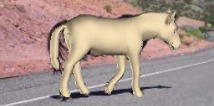} &

\includegraphics[width=0.3\linewidth]{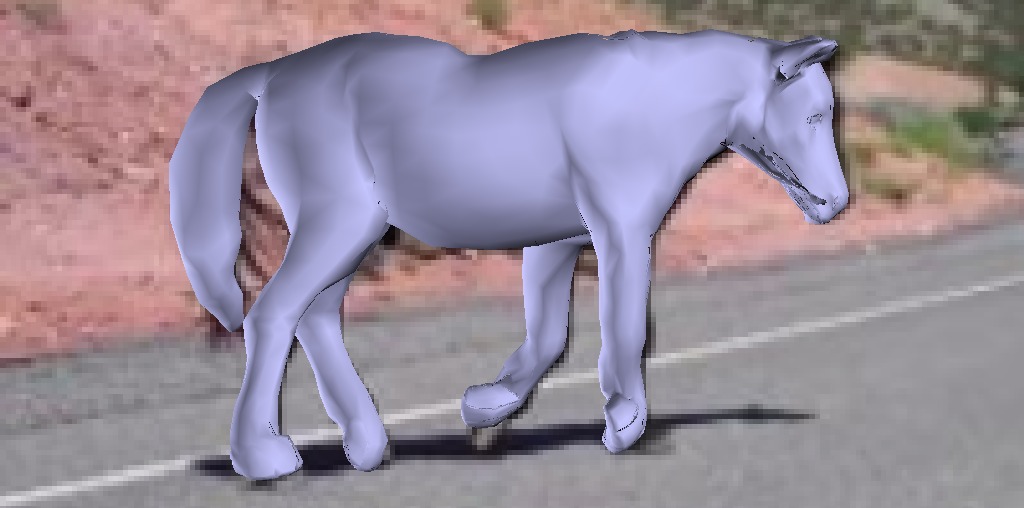} \\

RGB & SMAL \cite{zuffi2017menagerie} & \textbf{Ours}
\end{tabular}
\caption{Our results are comparable in quality to SMAL~\cite{zuffi2017menagerie}, but note that we do not require hand-clicked keypoints.
}
\label{fig:compar_smal}
\end{figure}

\subsection{Model fitting}
The predicted joint positions and silhouette are input to the optimization phase, which proceeds in four stages. The first stage solves for the model's global rotation and translation parameters, which positions the camera. We follow SMPLify~\cite{bogo2016keep} by solving this camera stage for torso points only, which remain largely fixed through shape and pose variation. We then solve for all shape, pose and translation parameters and gradually decrease the emphasis of the priors. The silhouette term is introduced in the penultimate stage, as otherwise we find this can lead to the optimizer finding unsatisfactory local minima.

\begin{figure}[h!]
\def\bb{\rule{2in}{0pt}\rule{0pt}{1in}}
\def\lp#1[#2]#3{\parbox{0.16\linewidth}{\labelledpic{#1}{\includegraphics[#2]{#3}}}}
\begin{tabular}{cccccc}
\includegraphics[trim={0 2cm 0 1.25cm},clip,width=0.16\linewidth]{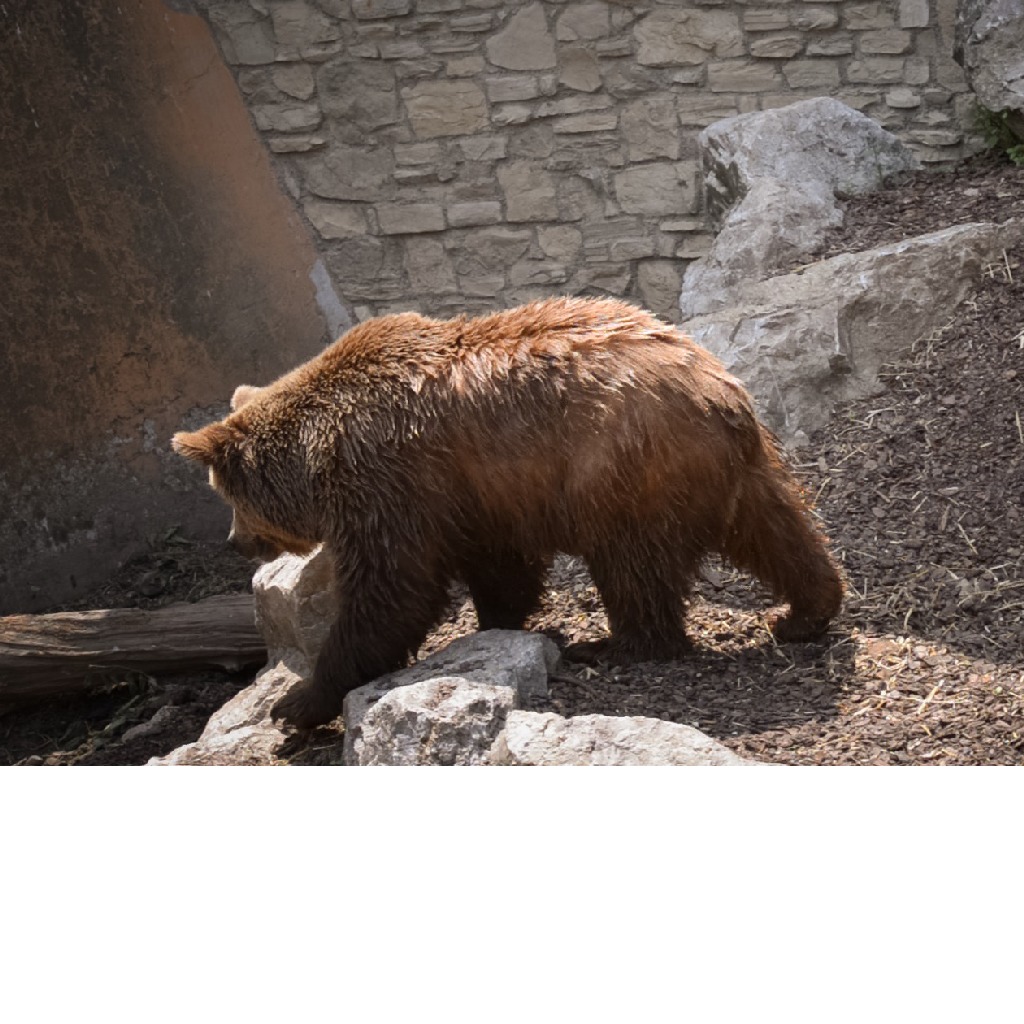} & 
\includegraphics[trim={0 2cm 0 1.25cm},clip,width=0.16\linewidth]{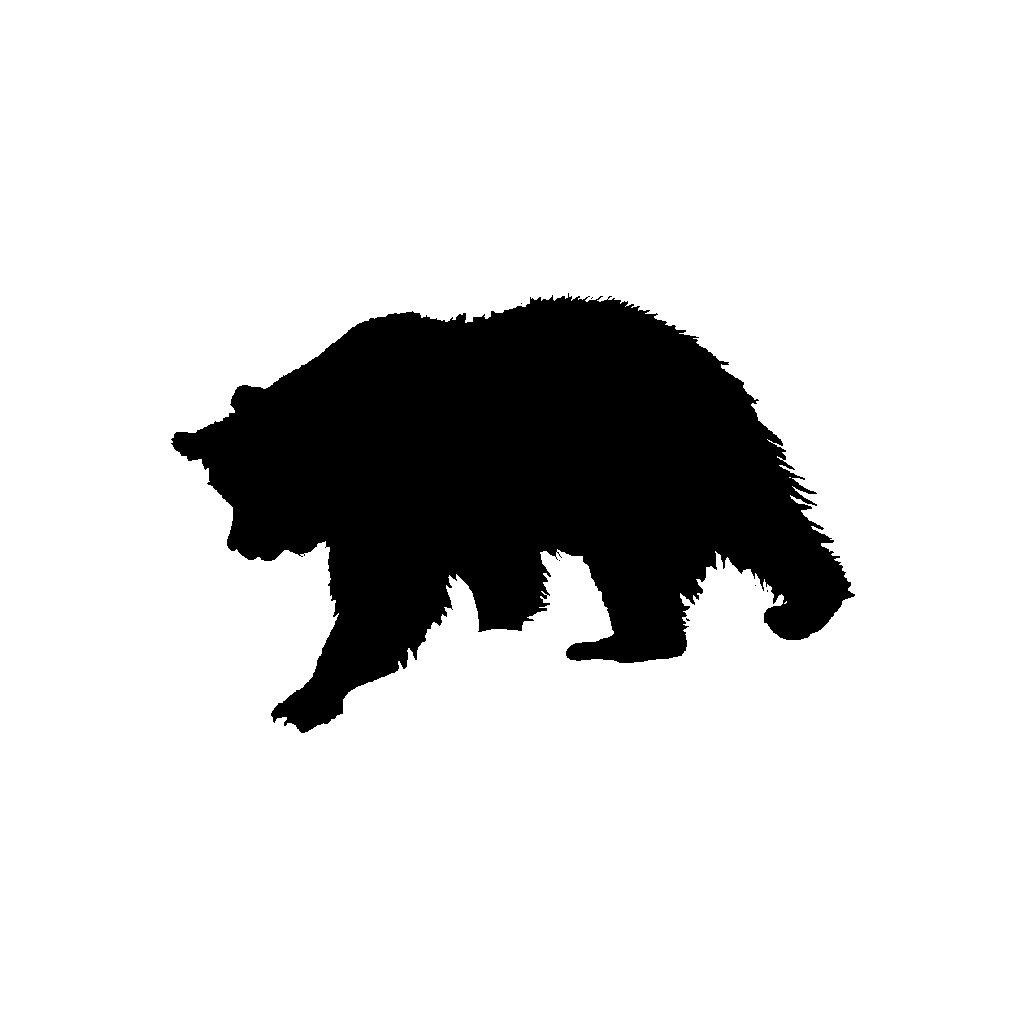} & 
\includegraphics[trim={0 2cm 0 1.25cm},clip,width=0.16\linewidth]{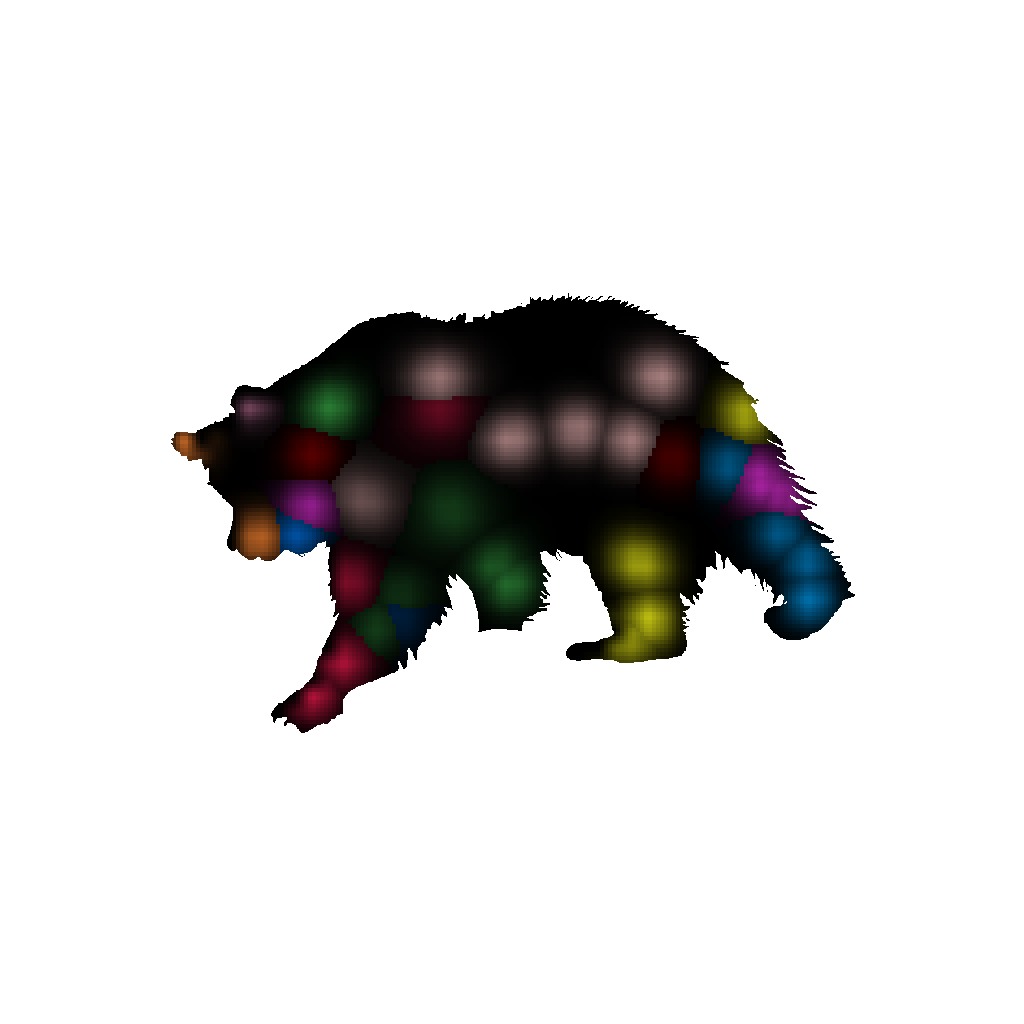} & 
\includegraphics[trim={0 2cm 0 1.25cm},clip,width=0.16\linewidth]{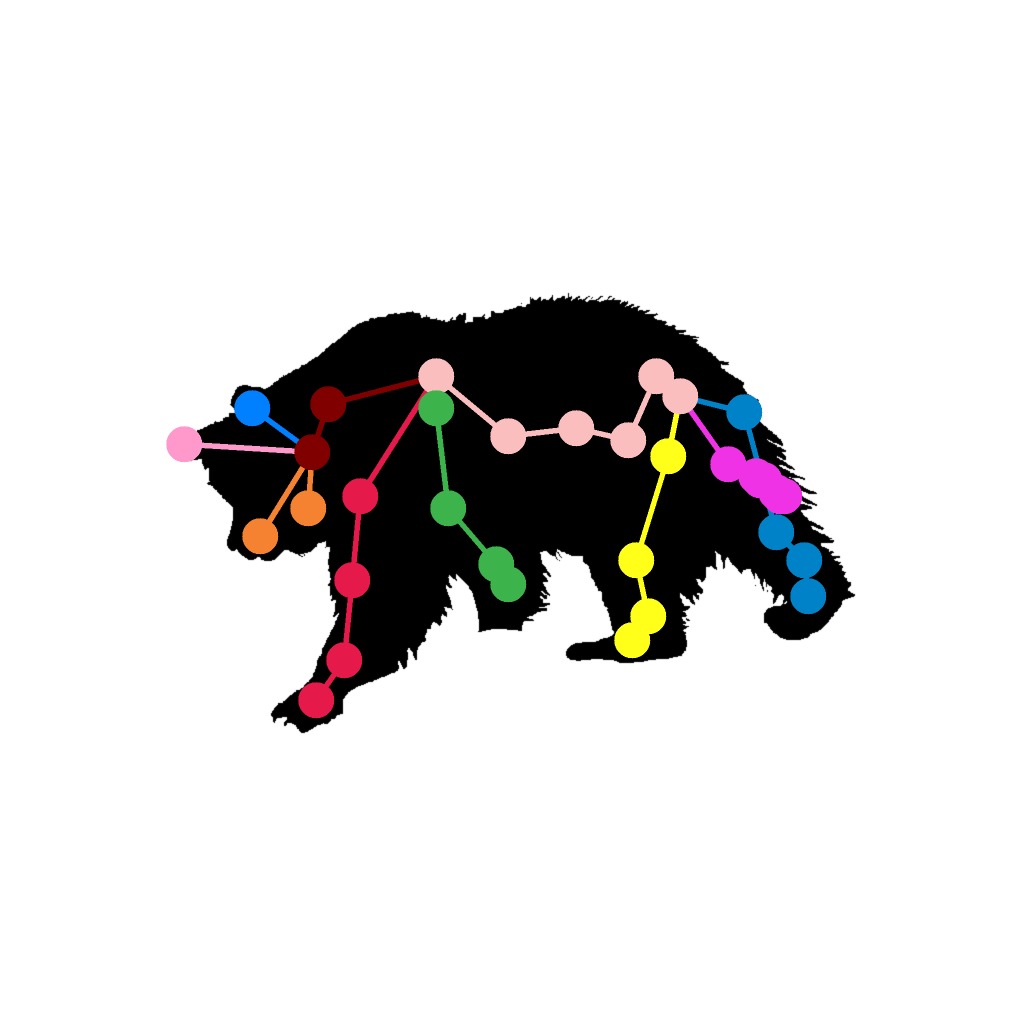} &
\includegraphics[trim={0 2cm 0 1.25cm},clip,width=0.16\linewidth]{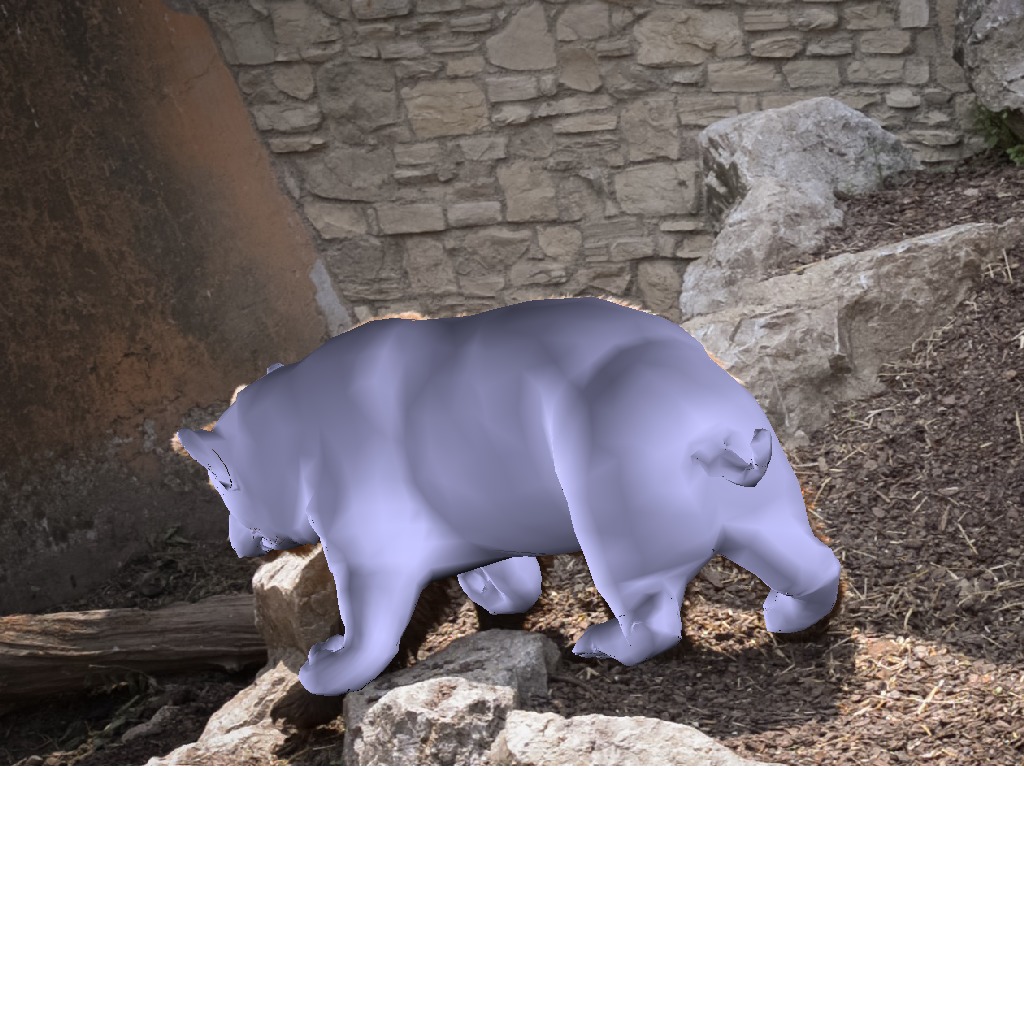} & 
\includegraphics[trim={0 2cm 0 1.25cm},clip,width=0.16\linewidth]{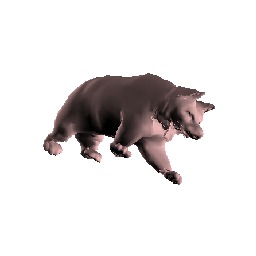} \\

\includegraphics[trim={0 0.5cm 0 0.5cm},clip,width=0.16\linewidth]{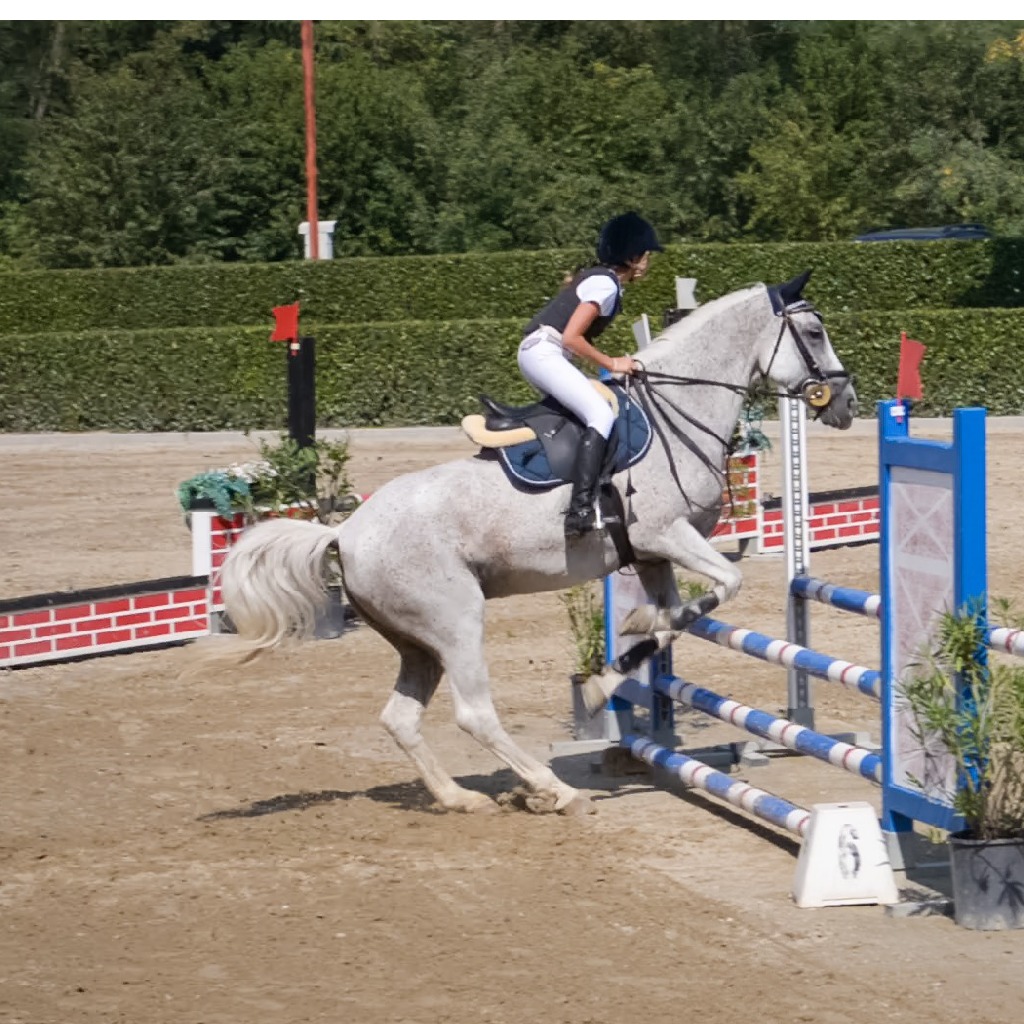} & 
\includegraphics[trim={0 0.5cm 0 0.5cm},clip,width=0.16\linewidth]{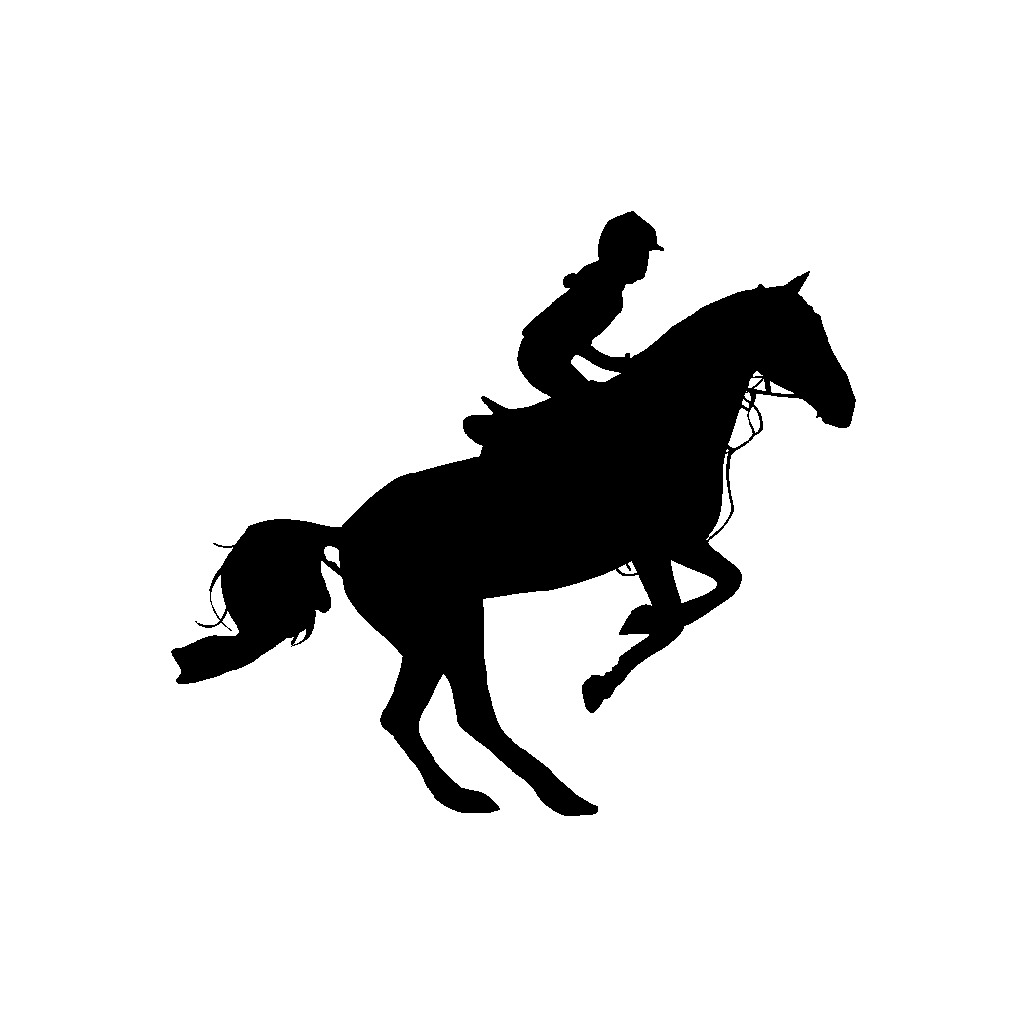} & 
\includegraphics[trim={0 0.5cm 0 0.5cm},clip,width=0.16\linewidth]{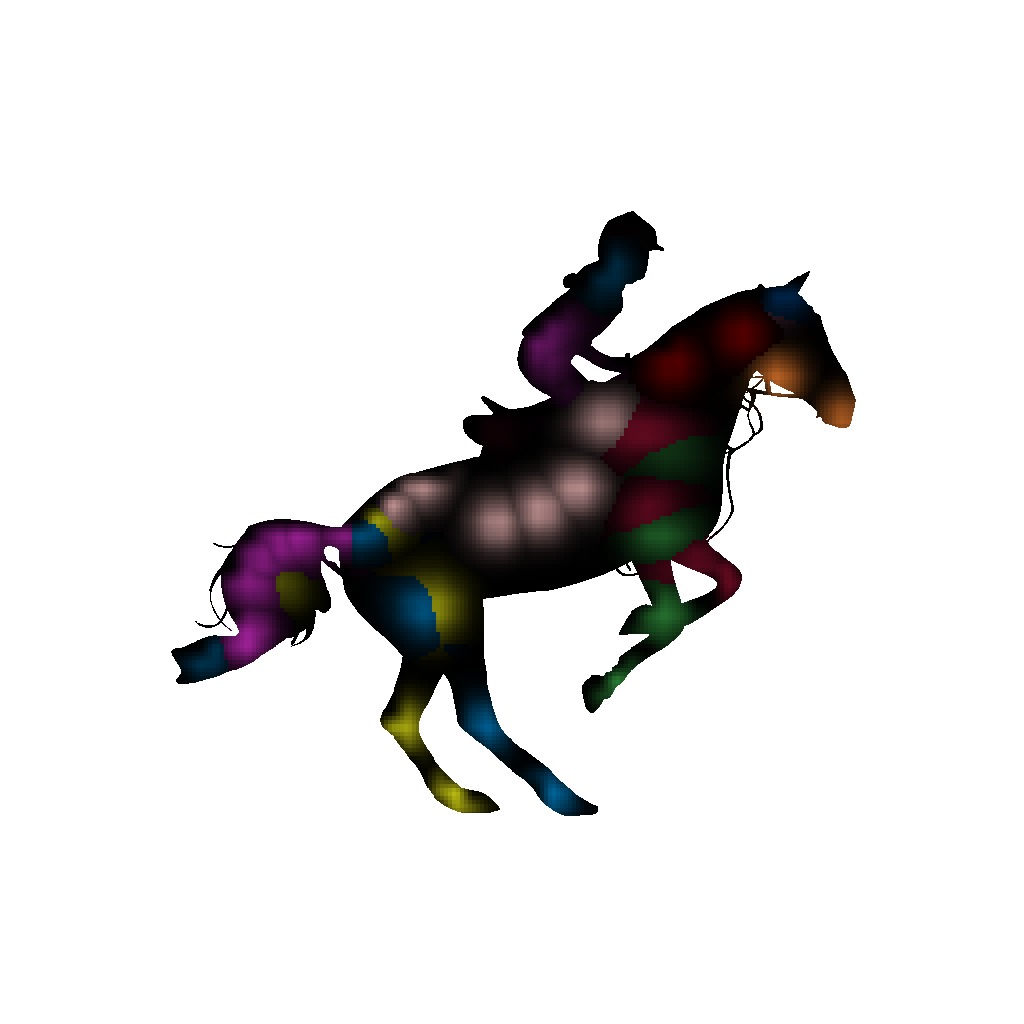} & 
\includegraphics[trim={0 0.5cm 0 0.5cm},clip,width=0.16\linewidth]{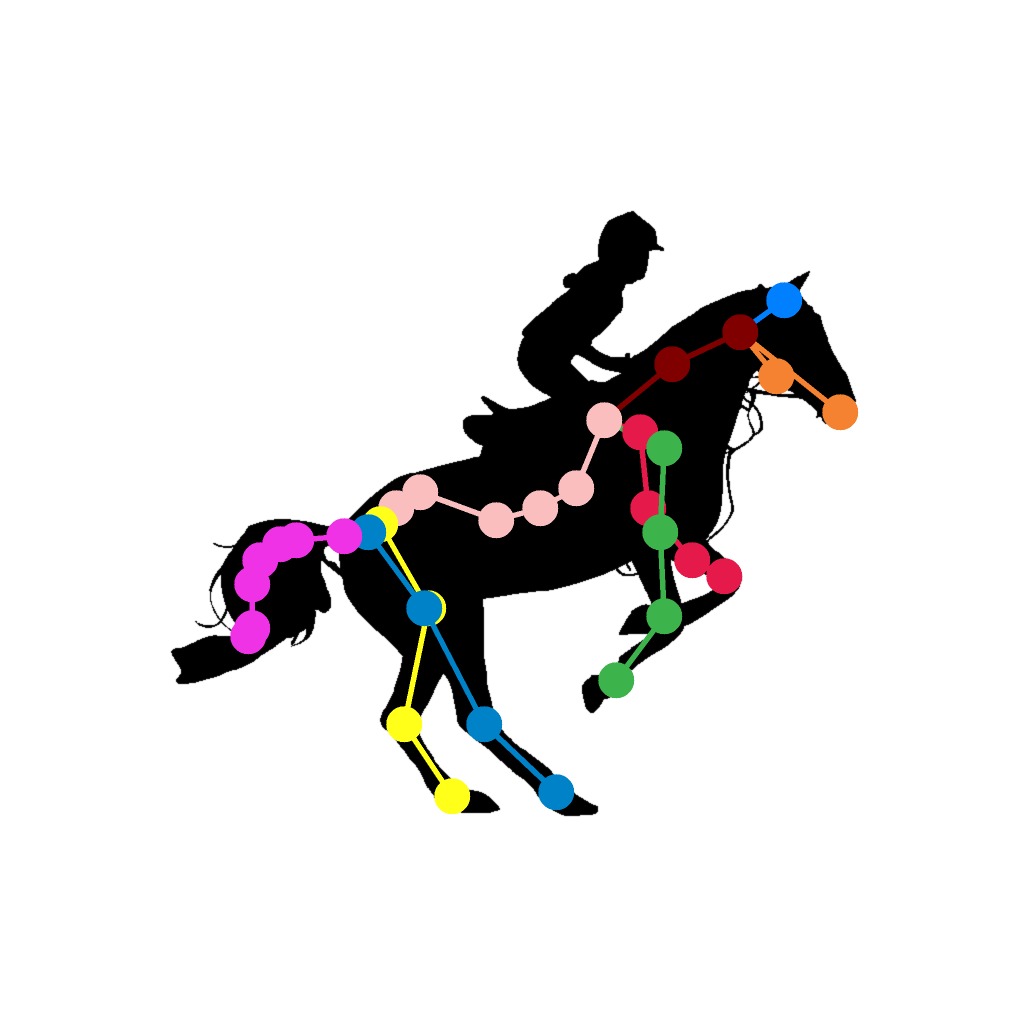} &
\includegraphics[trim={0 0.5cm 0 0.5cm},clip,width=0.16\linewidth]{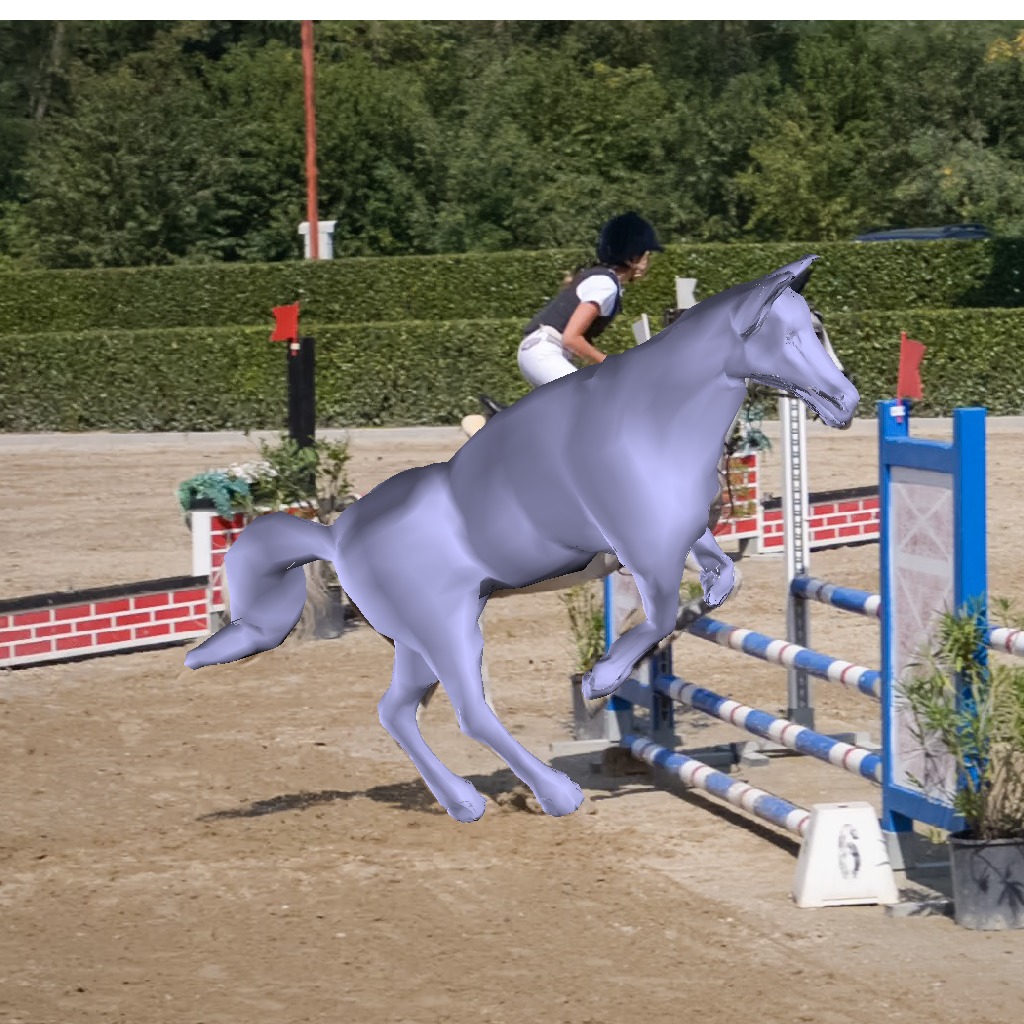} & 
\includegraphics[trim={0 0.5cm 0 0.5cm},clip,width=0.16\linewidth]{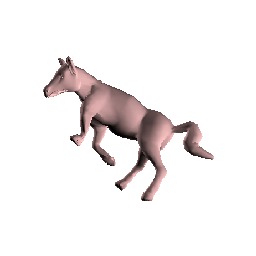} \\

\includegraphics[trim={0 1.5cm 0 1.5cm},clip,width=0.16\linewidth]{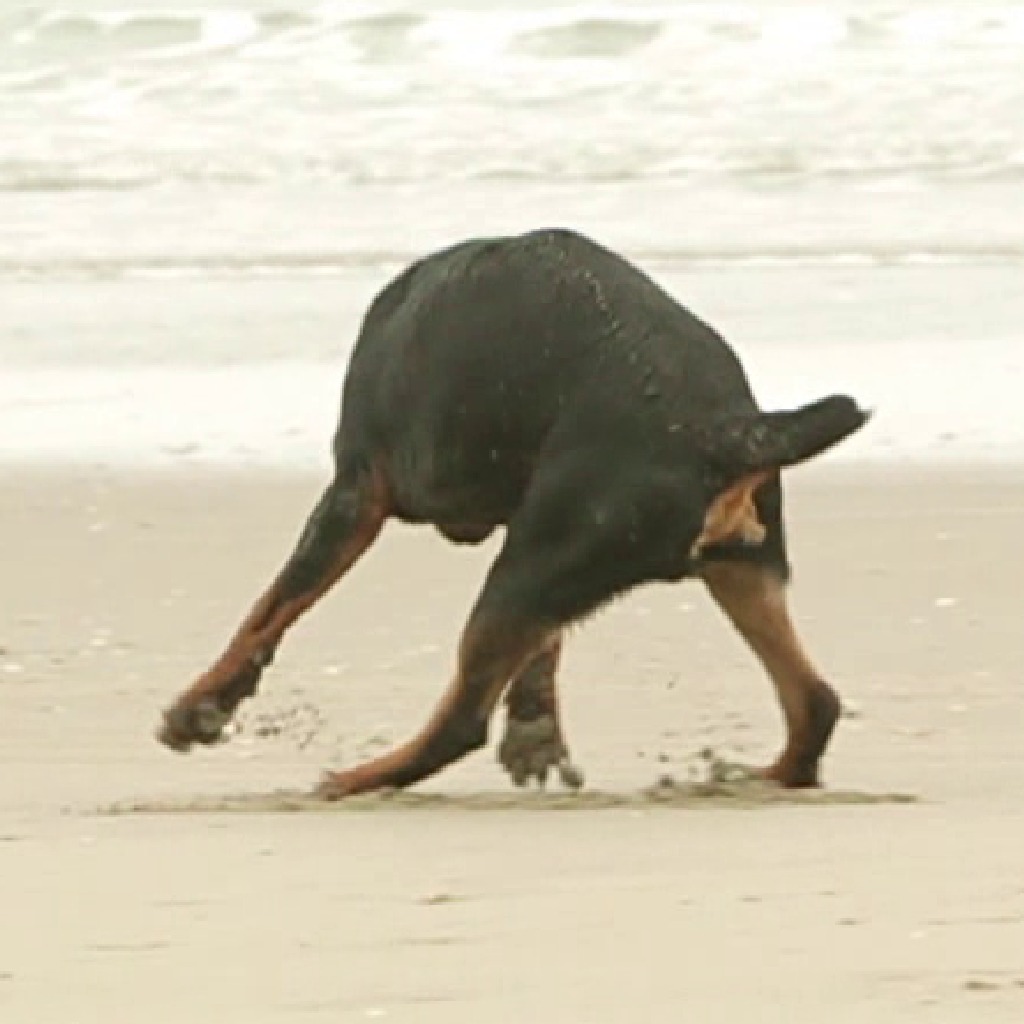} & 
\includegraphics[trim={0 1.5cm 0 1.5cm},clip,width=0.16\linewidth]{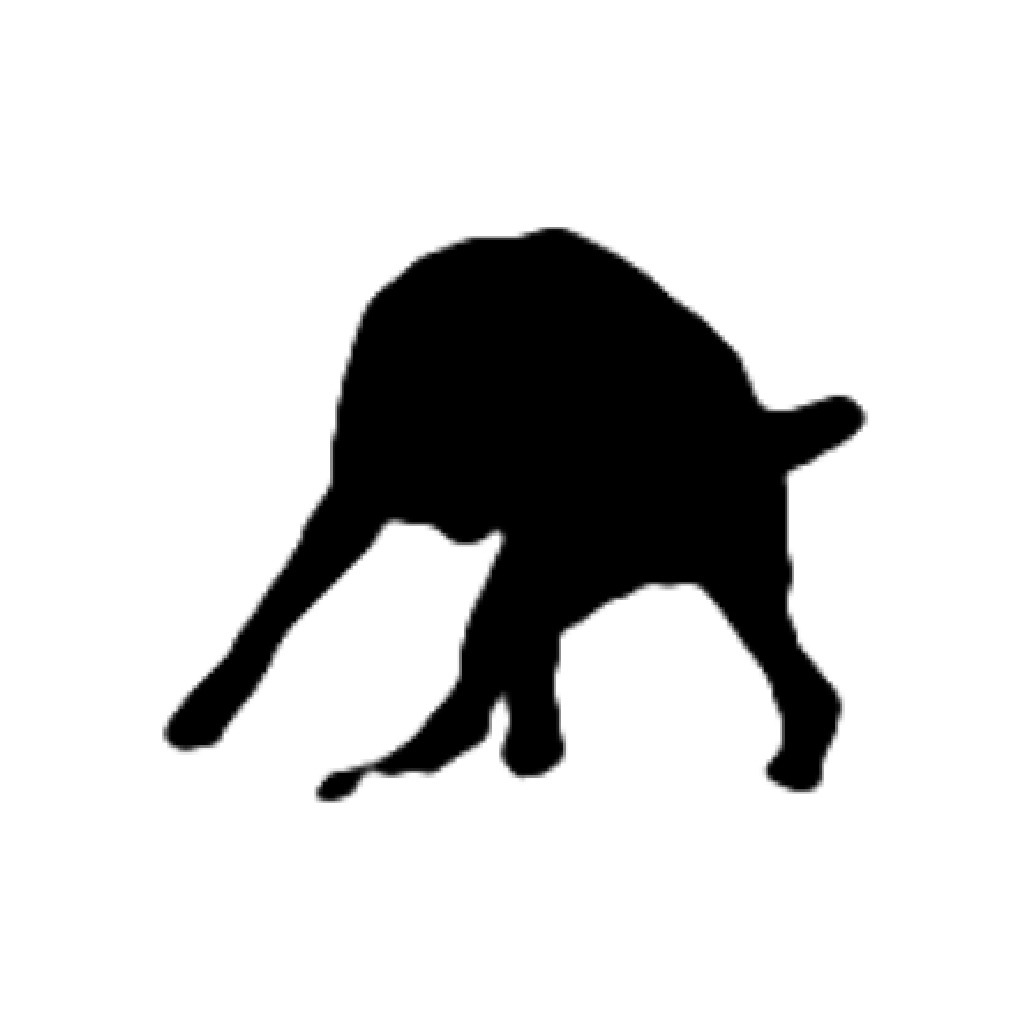} & 
\includegraphics[trim={0 1.5cm 0 1.5cm},clip,width=0.16\linewidth]{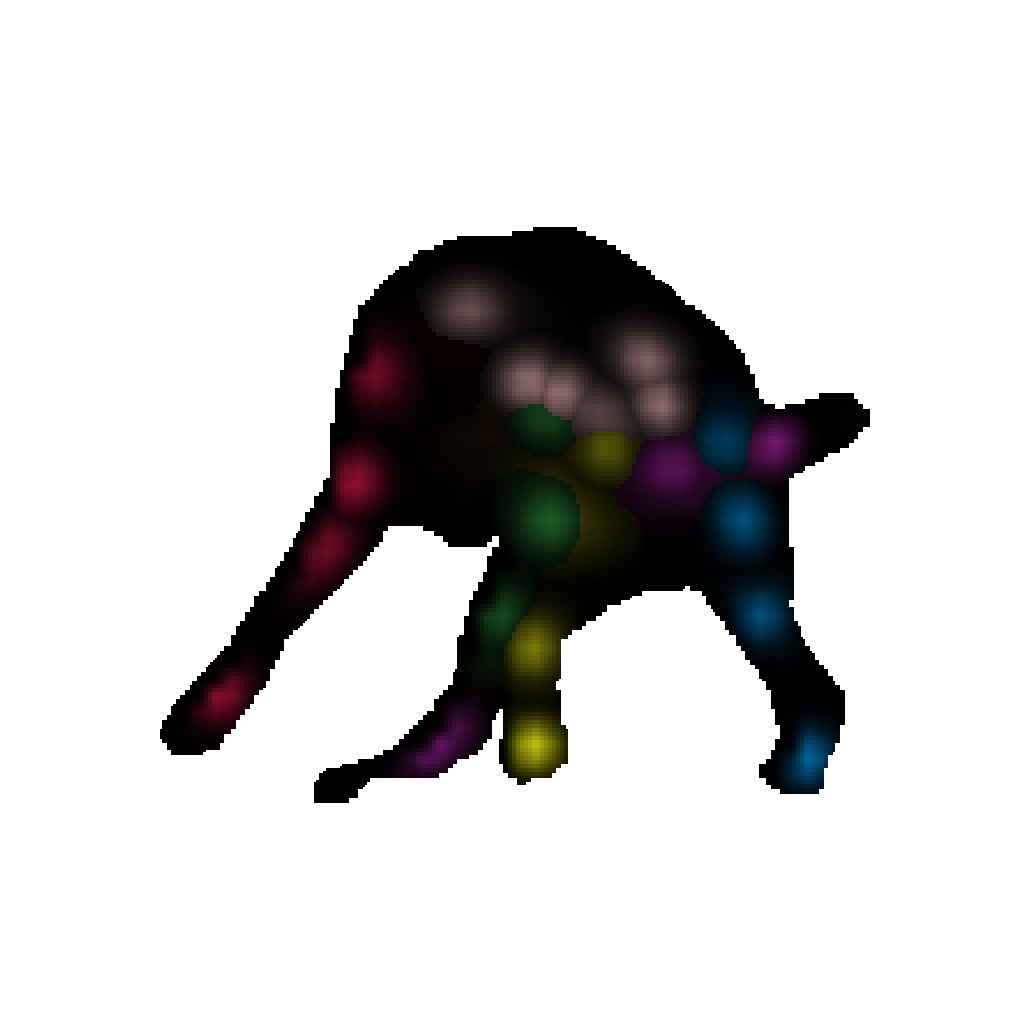} & 
\includegraphics[trim={0 1.5cm 0 1.5cm},clip,width=0.16\linewidth]{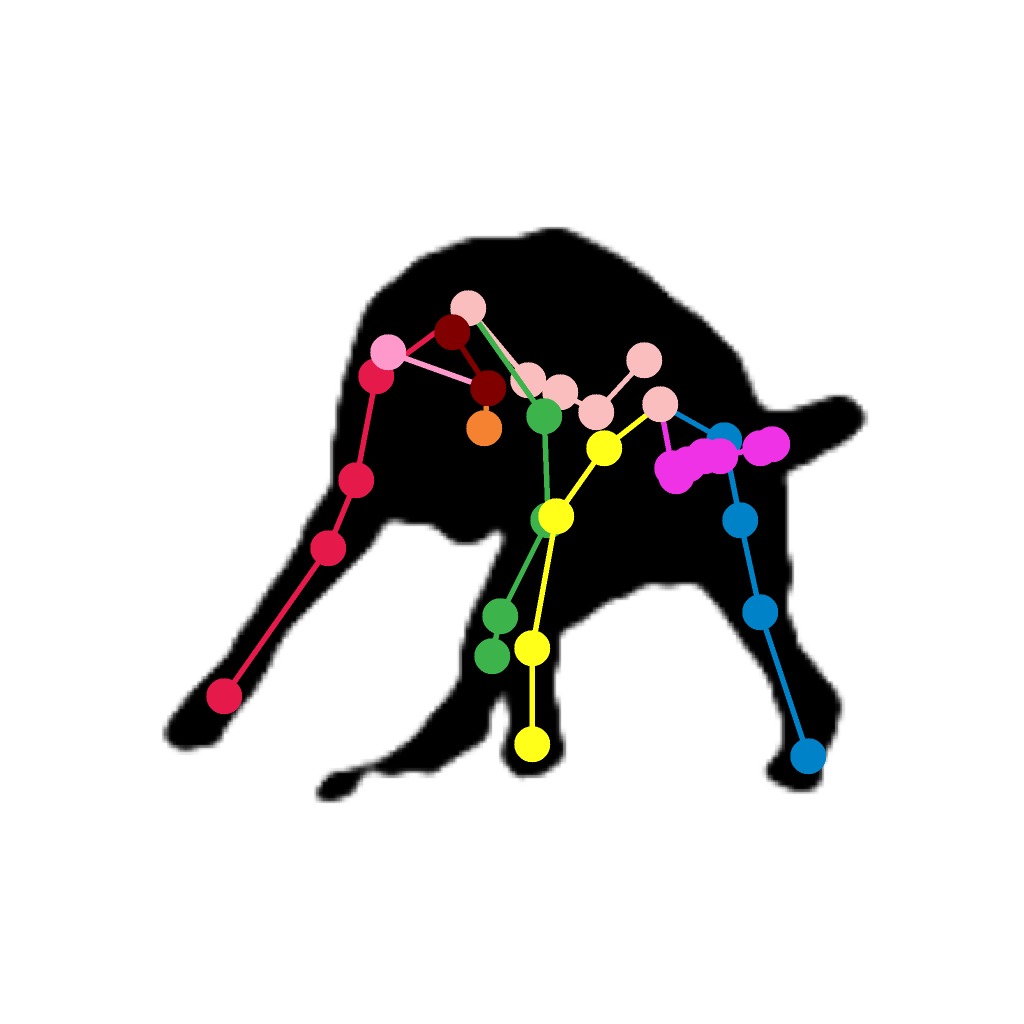} &
\includegraphics[trim={0 1.5cm 0 1.5cm},clip,width=0.16\linewidth]{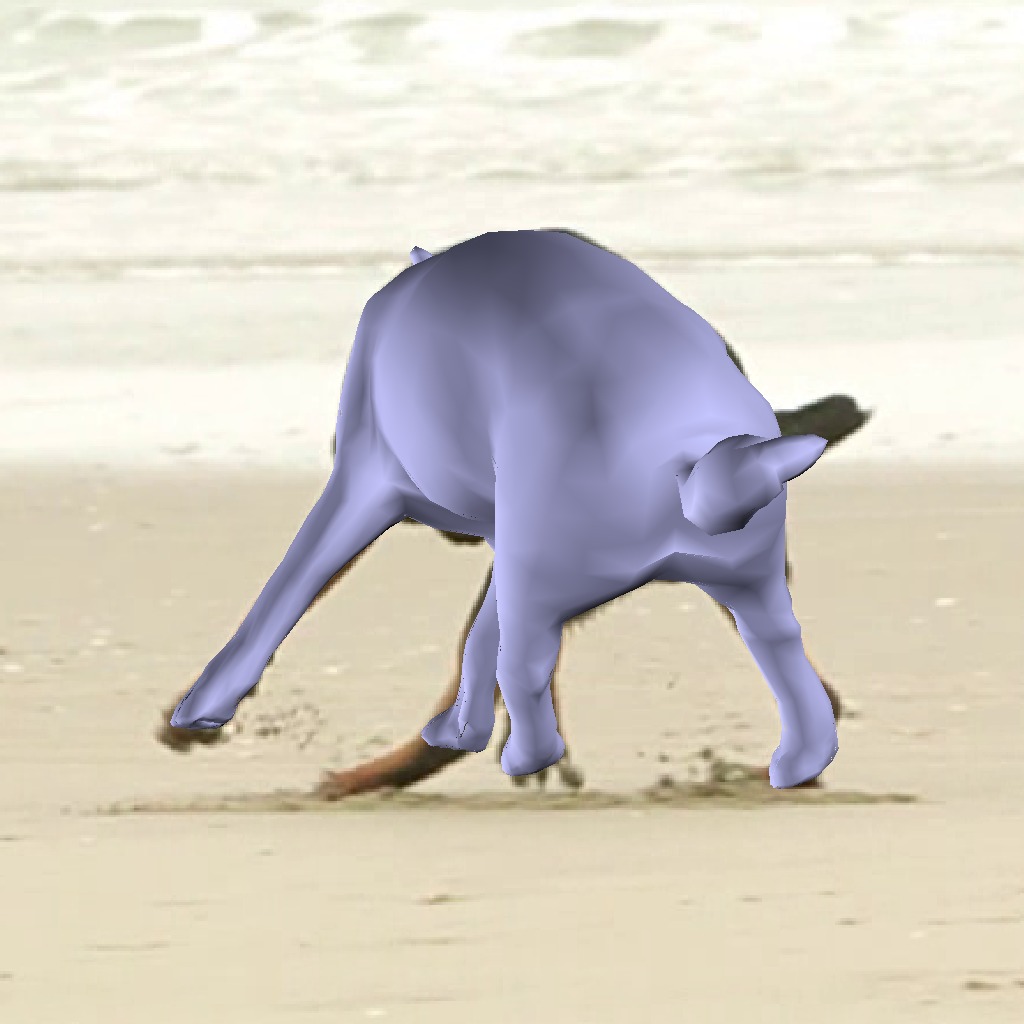} & 
\includegraphics[trim={0 1.5cm 0 1.5cm},clip,width=0.16\linewidth]{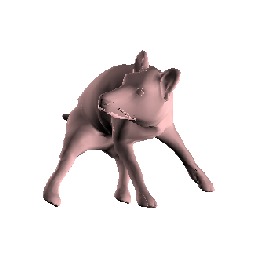} \\

\lp a[trim={0 1cm 0 1cm},clip,width=0.98\linewidth]{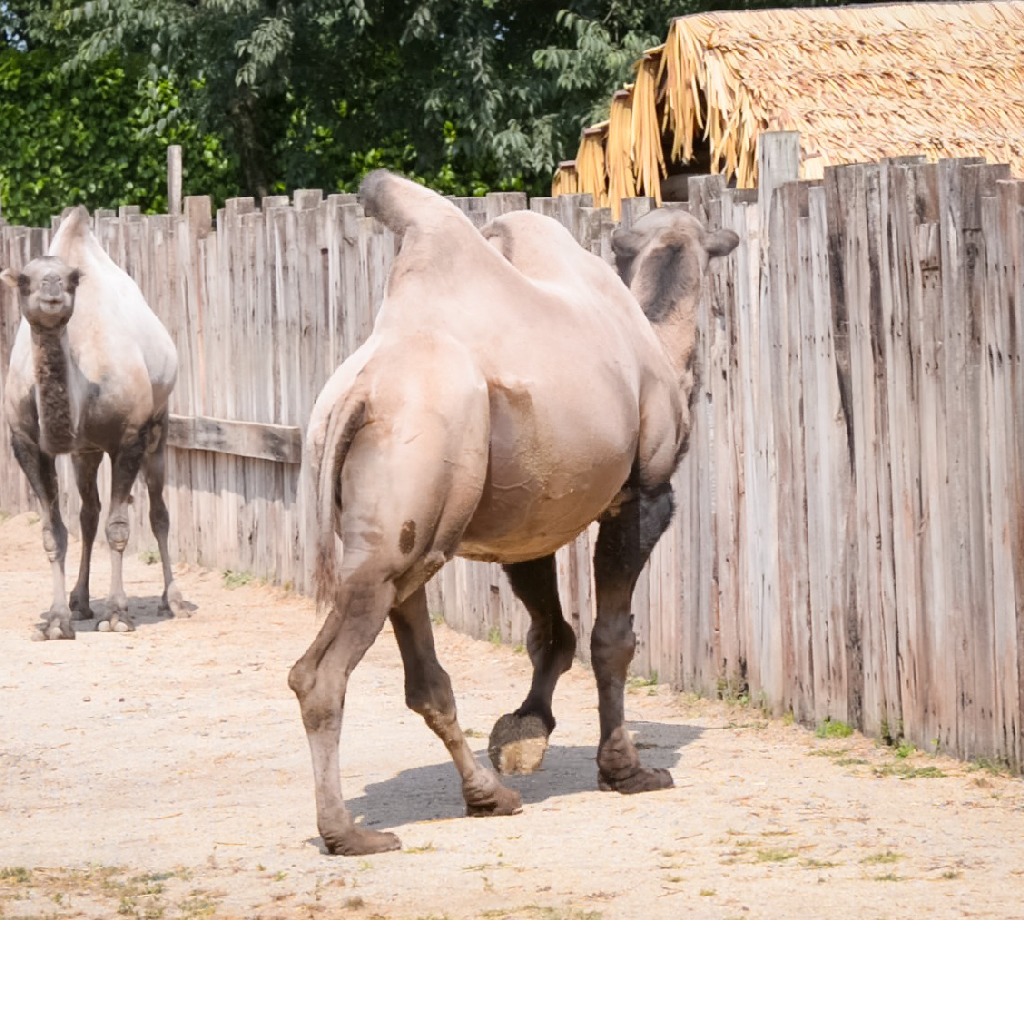} & 
\lp b[trim={0 1cm 0 1cm},clip,width=0.98\linewidth]{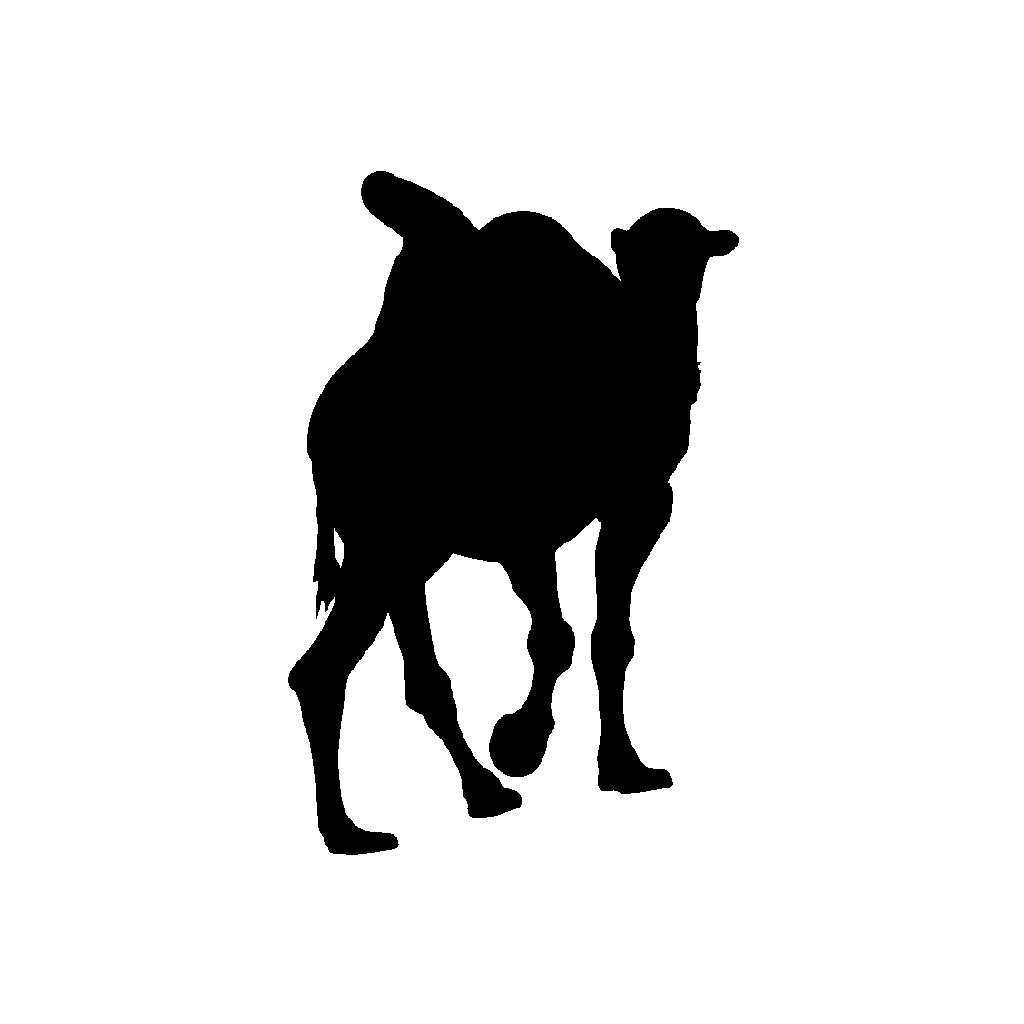} & 
\lp c[trim={0 1cm 0 1cm},clip,width=0.98\linewidth]{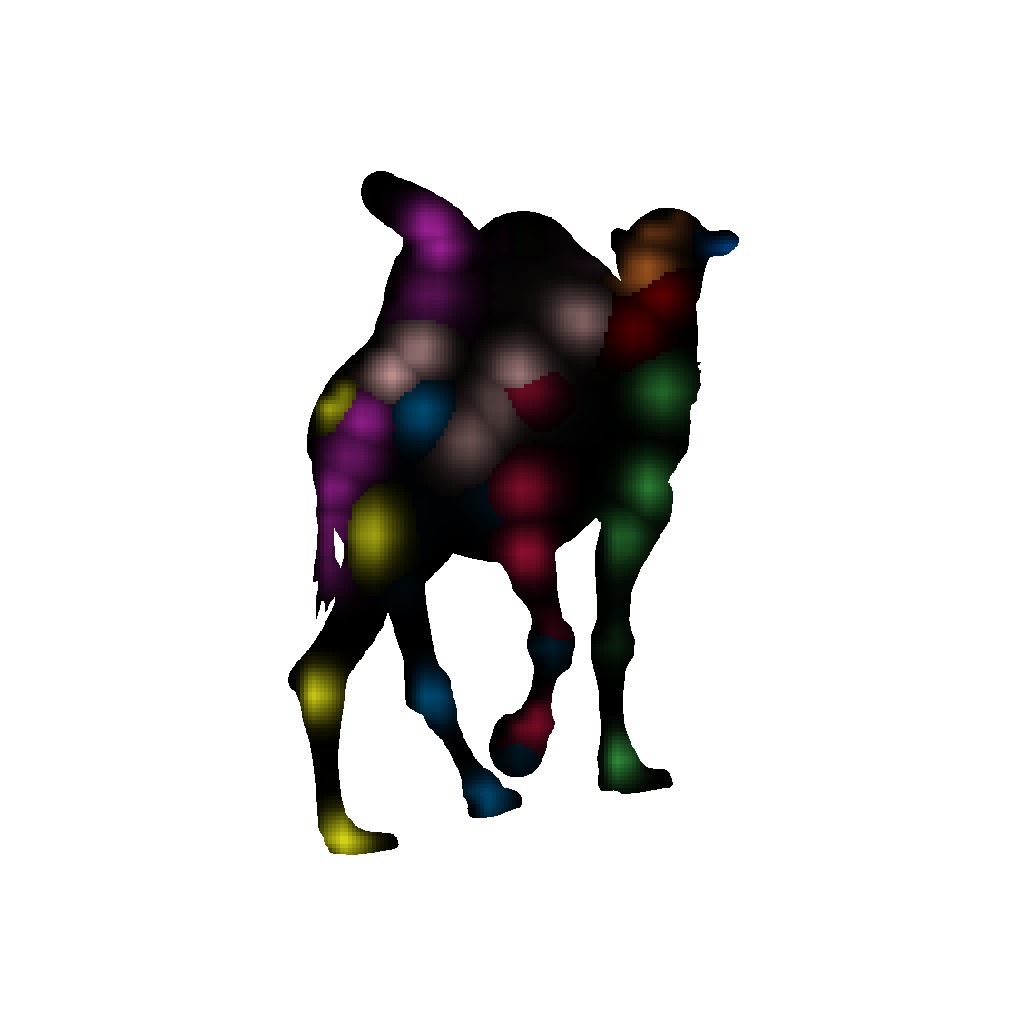} & 
\lp d[trim={0 1cm 0 1cm},clip,width=0.98\linewidth]{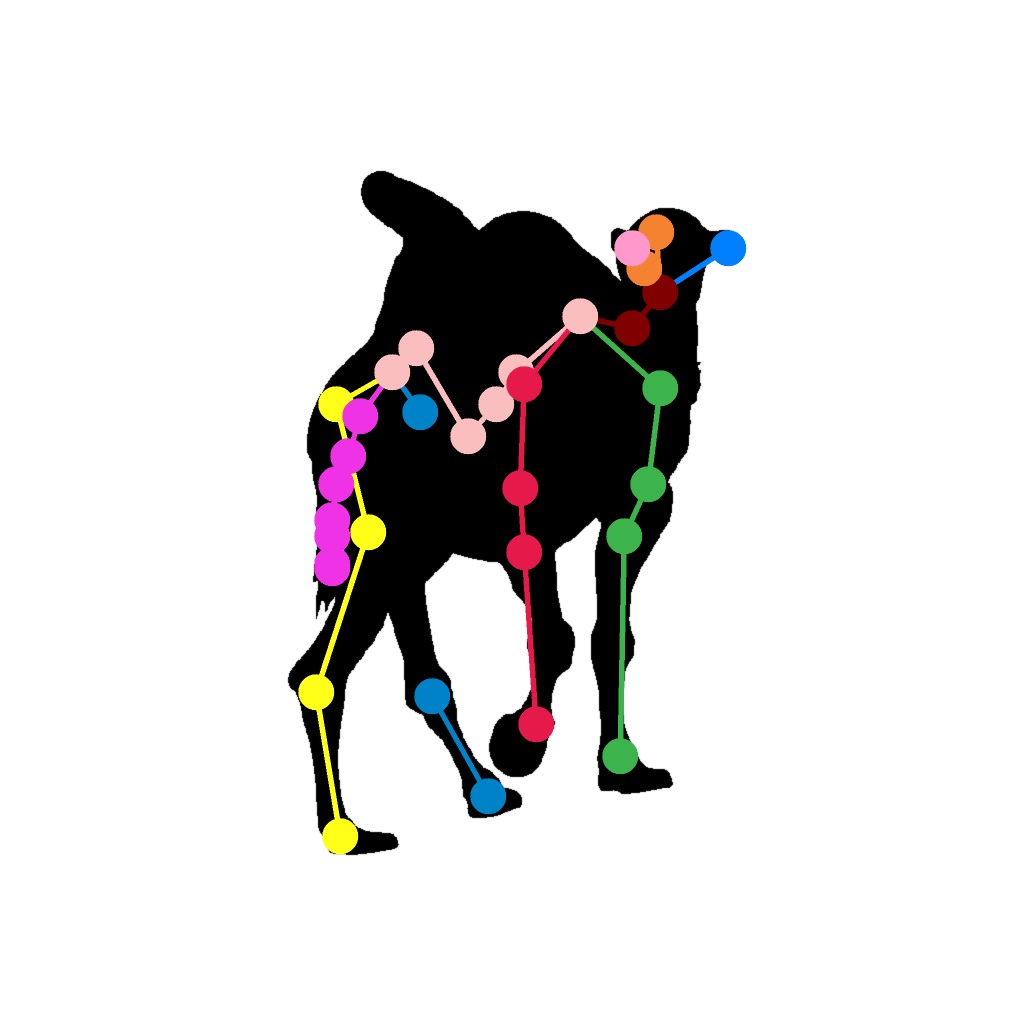} &
\lp e[trim={0 1cm 0 1cm},clip,width=0.98\linewidth]{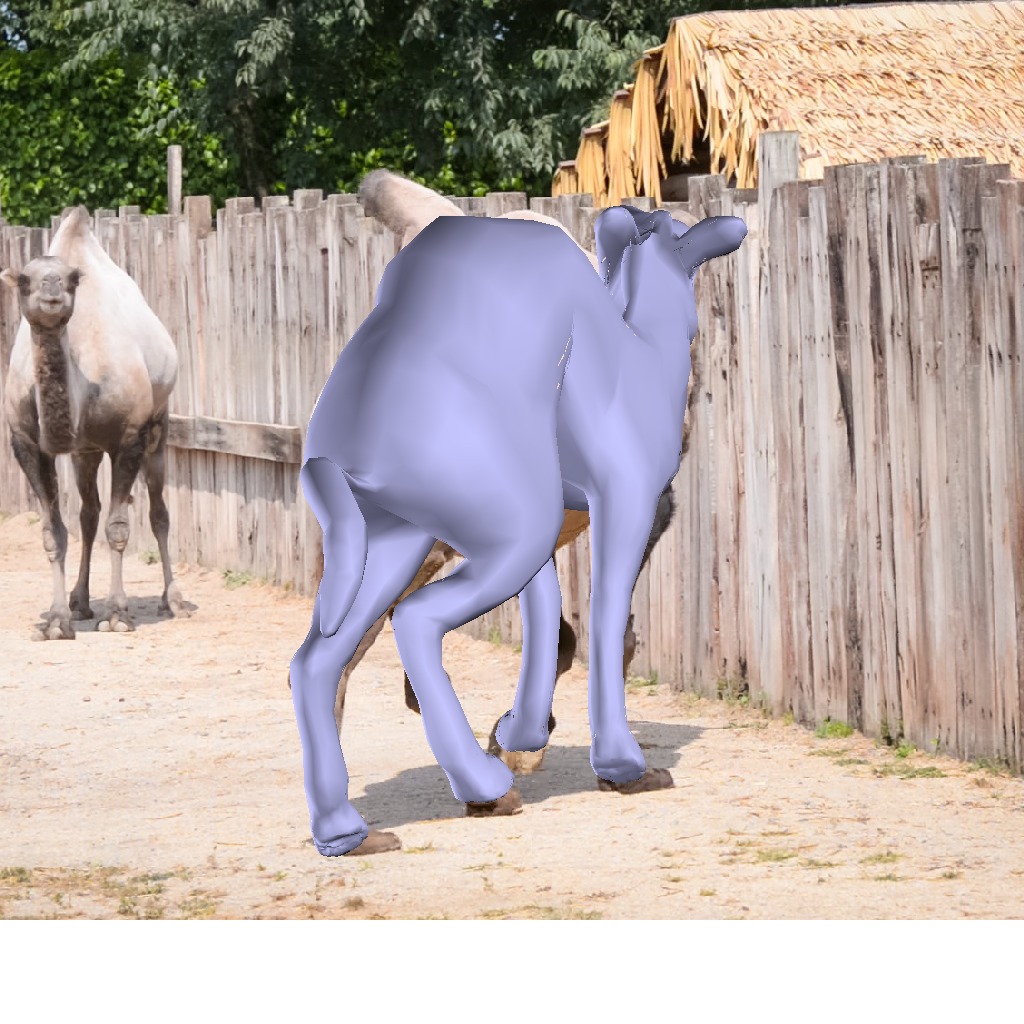} & 
\lp f[trim={0 1cm 0 1cm},clip,width=0.98\linewidth]{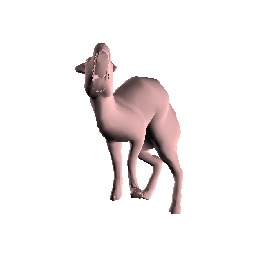} 
\end{tabular}
\caption{Example results on various animals. From left to right: RGB input, extracted silhouette, network-predicted heatmaps, OJA-processed joints, overlay 3D fit and alternative view.}
\label{fig:example_results}
\end{figure}

The final outputs of our optimization pipeline are shown in \figref{example_results}. In each of the cases illustrated the optimizer is able to successfully find a set of pose and shape parameters which, when rendered, closely resembles the input image. The final row of \figref{example_results} demonstrates the generalizability of the proposed method: the algorithm is able to find a reasonable pose despite no camel figurines being included in the original SMAL model.

\subsubsection*{Comparison to other work.} We compare our approach visually to that given by Zuffi {\em et al.}~\cite{zuffi2017menagerie}. Recall that their results require hand-clicked keypoints whereas ours fits to points predicted automatically by the hourglass network, which was trained on synthetic animal images. Further, their work is optimized for single frame fitting and is tested on animals in simple poses, whereas we instead focus on the more challenging task of tracking animals in video. \figref{compar_smal} shows the application of our model to a number of single frame examples from the SMAL result data~\cite{zuffi2017menagerie}.

\subsubsection*{Quantitative experiments.}
\begin{figure}[t!]
\begin{tabular}{cccccc}
\includegraphics[width=0.16\linewidth]{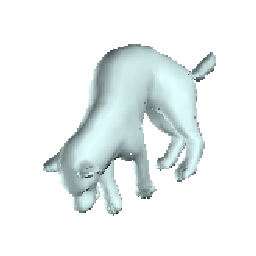} & 
\includegraphics[width=0.16\linewidth]{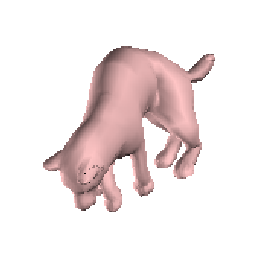} &

\includegraphics[width=0.16\linewidth]{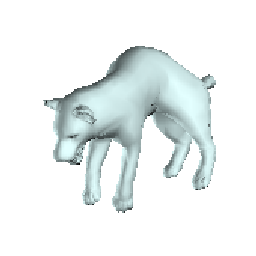} &
\includegraphics[width=0.16\linewidth]{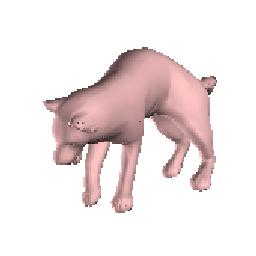} &

\includegraphics[width=0.16\linewidth]{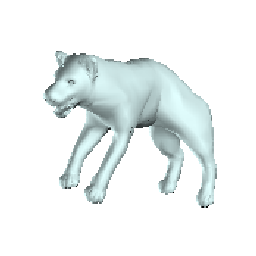} &
\includegraphics[width=0.16\linewidth]{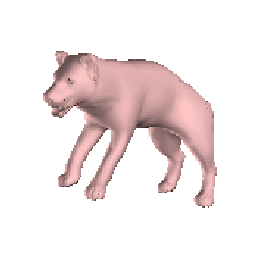} \\
\end{tabular}
\caption{Evaluating synthetic data. Green models: ground truth, Orange models: predicted. Frames 5, 10 and 15 of sequence 4 shown. Error on this sequence 22.9.}
\label{fig:synth}
\end{figure}

There is no existing ground truth dataset for comparing reconstructed 3D animal meshes, but an estimate of quantitative error is obtained by testing on synthetic sequences for a range of quadruped species. These are generated by randomly deforming the model and varying the camera position to animate animal motion, see Figure~\ref{fig:synth}. Table~\ref{tab:synthetic} shows results on these sequences. 

\subsection{Automatic silhouette prediction}
While not the main focus of our work, we are able to perform the full 3D reconstruction process from an input image with no user intervention. We achieve this by using the DeepLabv3+ network~\cite{deeplabv3plus} as a front-end segmentation engine to automatically generate animal silhouettes. This network was trained on the PASCAL VOC 2012 dataset, which includes a variety of animal quadruped classes. An example result generated using the fully automatic pipeline is shown in \figref{overview}.

\clearpage

\vspace{-3em}

\vspace{-2em}
\section{Conclusions}
In this work we have introduced a technique for 3D animal reconstruction from video using a quadruped model parameterized in shape and pose. By incorporating automatic segmentation tools, we demonstrated that this can be achieved with no human intervention or prior knowledge of the species of animal being considered. Our method performs well on examples encountered in the real world, generalizes to unseen animal species and is robust to challenging input conditions.

\section{Acknowlegements}
The authors would like to thank Ignas Budvytis and James Charles for technical discussions, Peter Grandi and Raf Czlonka for their impassioned IT support, the Biggs' family for the excellent title pun and Philippa Liggins for proof reading.
\clearpage
\bibliographystyle{splncs}
\bibliography{egbib}

\end{document}